\documentclass{article}

\usepackage{microtype}
\usepackage{graphicx}
\usepackage{subfigure}
\usepackage{booktabs} 

\usepackage{hyperref}
\usepackage{microtype}
\usepackage{graphicx}
\usepackage{subfigure}
\usepackage{booktabs} 
\usepackage{amsmath,amssymb,amsthm,bbm,mathtools,bm}

\newtheorem{counter}{Counter}[section]
\newtheorem{lem}[counter]{Lemma}
\newtheorem{defn}[counter]{Definition}
\newtheorem{thm}[counter]{Theorem}
\newtheorem{rmk}[counter]{Remark}

\newcommand{\real}{\mathbb{R}}

\DeclareMathOperator{\grad}{grad}
\DeclareMathOperator{\diver}{div}
\DeclareMathOperator{\curl}{curl}
\DeclareMathOperator{\relu}{ReLU}
\DeclareMathOperator{\tr}{tr}

\usepackage[accepted]{icml2021}

\icmltitlerunning{DAGs with No Curl}

\begin{document}

\twocolumn[
\icmltitle{DAGs with No Curl: An Efficient DAG Structure Learning Approach}



\icmlsetsymbol{equal}{*}

\begin{icmlauthorlist}
\icmlauthor{Yue Yu}{to}
\icmlauthor{Tian Gao}{ibm}
\icmlauthor{Naiyu Yin}{rpi}
\icmlauthor{Qiang Ji}{rpi}
\end{icmlauthorlist}

\icmlaffiliation{to}{Department of Mathematics, Lehigh University, Bethlehem, PA.}
\icmlaffiliation{ibm}{IBM Research, Yorktown Heights, NY.}
\icmlaffiliation{rpi}{Department of Electrical, Computer, and Systems Engineering,
Rensselaer Polytechnic Institute,
Troy, NY.}

\icmlcorrespondingauthor{Yue Yu}{yuy214@lehigh.edu}

\icmlkeywords{Machine Learning, ICML}

\vskip 0.3in
]



\printAffiliationsAndNotice{}  

\begin{abstract}
Recently directed acyclic graph (DAG) structure learning is formulated as a constrained continuous optimization problem with continuous acyclicity constraints and was solved iteratively through subproblem optimization. To further improve efficiency, we propose a novel learning framework to model and learn the weighted adjacency matrices in the DAG space directly. Specifically, we first show that the set of weighted adjacency matrices of DAGs are equivalent to the set of weighted gradients of graph potential functions, and one may perform structure learning by searching in this equivalent set of DAGs. To instantiate this idea,  we propose a new algorithm,  DAG-NoCurl, which  solves the optimization problem efficiently with a two-step procedure: $1)$ first we find an initial cyclic solution to the optimization problem, and $2)$ then we employ the Hodge decomposition of graphs and learn an acyclic graph  by projecting the cyclic graph to the gradient of a potential function.  Experimental studies on benchmark datasets demonstrate that our method provides comparable accuracy but better efficiency than baseline DAG structure learning methods on both linear and generalized structural equation models, often by more than one order of magnitude.
\end{abstract}

\section{Introduction}

Bayesian Networks (BN) have been widely used in various machine learning applications~\citep{ott2004finding,glymour1999computation}. Efficient structure learning of BN remains an active area of research. The structure takes the form of a directed acyclic graph (DAG) and plays a vital part in other machine learning sub-areas such as causal inference~\citep{pearl88}. However, DAG learning is proven to be  NP-hard~\citep{chickering2004large} and scalability becomes a major issue.

Conventional DAG learning methods  usually use independence tests  \cite{spirtes2000causation, MMPCcor} or perform  score-and-search for discrete variables, with different scoring functions \citep{huang2018generalized} or search procedures \citep{silander06, chickering02,cussens11}. Learning DAG structures for continuous variables is often limited to Gaussian  models \citep{yuan2007model, foygel2010extended, schmidt2007learning, mohan2012structured, mohammadi2015bayesian}. Recently, a fully continuous optimization formulation is proposed \citep{zheng2018dags}, which transforms the discrete DAG constraint into a continuous equality constraint. This approach enables a suite of continuous optimization techniques such as gradient descent to be used, and  has been extend to a more general parameter class with various neural methods \citep{yu2019dag,lachapelle2019gradient,zhu2019causal,ng2019masked}. 

In this work, we  take a step further and investigate if we could directly optimize in the DAG space, hence resulting in a structure learning approach \textit{without any explicit} DAG constraint.  To this end, we propose a continuous optimization framework for DAG structure learning, which \textit{implicitly} enforces the acyclicity of the learned graph. 
\citet{varando2020learning} and \citet{ng2020role} have studied the usage of empirical correlation  and covariance matrices for similar purposes. Different from their works, we propose a new graph-exterior-calculus-based framework of DAG such that one can  directly optimize in the DAG space. To  solve the resultant unconstrained optimization problem, we further propose  an efficient algorithm, DAG-NoCurl, developed based on the graph Hodge theory \citep{jiang2011statistical}. Hodge theory \citep{hodge1989theory}, along with the related Helmholtz-Hodge Decomposition \citep{bhatia2012helmholtz} on vector fields, 
describes the decomposition of a vector field into  divergence-free and curl-free components. 
Hodge theory on graphs \citep{lim2015hodge} shows that a DAG is a sum of three components: a curl-free, a divergence-free, and a harmonic component, where  the curl-free component is an acyclic graph and hence motivates the naming of our algorithm.

DAG-NoCurl relies on the key step of mapping the adjacency weighted matrix of a directed graph onto its curl-free component, i.e., an acyclic graph.
The main advantages of the proposed method over conventional structure learning algorithms are: $1)$ the new model provides an equivalent representation of DAGs, which can be readily combined with many existing structural equation models \citep{zheng2018dags,yu2019dag}, $2)$ the new model naturally transfers the DAG structure learning problem as a continuous optimization framework, which avoids the combinatorial search and enables a suite of continuous optimization techniques; $3)$ comparing with other fully continuous optimization frameworks for DAG learning \citep{lachapelle2019gradient,yu2019dag,zheng2020learning}, our framework needs no explicit DAG constraints, many iterations, 
nor expensive post-processing, hence our learning approach achieves a substantially better computational efficiency.

\noindent\textbf{Contributions.} We make several major contributions in this work. $1)$ We propose a new model for DAGs based on the graph combinatorial gradient operator.  Specifically, we theoretically show that the weighted adjacency matrix of a DAG can be represented as the Hadamard product of a skew-symmetric matrix and the gradient of a potential function on graph vertices, and vice versa. 
$2)$ Based on the new model, we develop a new continuous optimization framework for DAG structure learning without any explicit constraint. 
$3)$ To  solve for the optimization problem efficiently, 
we propose a new DAG structure learning algorithm, DAG-NoCurl, that learns a weighted adjacency matrix to the original problem efficiently without constraints or iterations.  $4)$ We demonstrate the effectiveness of the proposed method on synthetic and benchmark datasets. 
While the optimization problem remains nonconvex, the new formulation can substantially improve the
efficiency,  often by more than one order of magnitude, while preserving the accuracy.


\section{Problem Statement and Motivation}

Let ${V}$ denote a set of $d$ numbers of random variables, $X=(X_1,\cdots,X_d)\in\real^d$ be an observation on ${V}$, and $\mathbb{D}$ denotes the space of DAGs $\mathcal{G}=(V,E)$ on $V$, we aim to learn a DAG $\mathcal{G}\in\mathbb{D}$ given $n$ i.i.d. observations of the random vector $X^i\in \real^d$, $i=1,\cdots,n$. We assume no latent variables.
To model $X$, we consider a (generalized) structural equation model (SEM) defined by a weighted adjacency matrix $A=[a_1|\cdots|a_d]\in \real^{d\times d}$ such that $\mathbb{E}(X_j|X_{pa(X_j)})=f(a_j^TX)$, where $pa(X_j)$ denote the parents of random variable $j$ in $V$. Therefore $[A]_{ij}\neq 0$ indicates a directed edge from vertex $i$ to vertex $j$ in a directed graph $\mathcal{G}_A$ and zero otherwise. With a slight abuse of notation, we will treat the weighted adjacency matrix $A$ as a (weighted) directed graph $\mathcal{G}_A$.


Given the data matrix $\mathbf{X}=(X^1,\cdots,X^n)\in\real^{d\times n}$ and a loss function $F(A,\mathbf{X})$ that measures the goodness of fit of $\mathbf{X}$ for $A$, in the \emph{structure learning} problem we aim to find the best DAG $A^*$ that minimizes $F(A, \mathbf{X})$. Hence, the overall objective can be written as: 
\begin{equation}\label{eqn:original}
\begin{aligned}
&&A^* & =  \underset{A}{\text{argmin}}
&&  F(A, \mathbf{X}) \\
&& &\text{subject to}
&& \mathcal{G}_A \in \mathbb{D} \quad \text{or} \quad h(A) = 0, 
\end{aligned}
\end{equation}
where $\mathbb{D}$ is the DAG space and  $h(A) = 0$ is an alternative continuous DAG constraint  \citep{zheng2018dags,wei2020dags}. Formally, $h(A)=\text{tr}(\exp(A\circ A))-d$ \cite{zheng2018dags} or $ h(A)=\text{tr}[(I + A\circ A/d)^d] -d$ \cite{yu2019dag}.

As one may know, the DAG space $\mathbb{D}$ for $d$ variables is super-exponential. 
For structure learning over discrete variables, many exact and approximate algorithms from observational data have been proposed ~\citep{PCalgorithm,campos2009,shimizu2014lingam} with different search strategies, such as dynamic programming \citep{smDP,koivisto2004exact}, A* \citep{Yuan13learning}, or integer programming \citep{jaakkola2010,cussens2016polyhedral}. In large-scale problems, approximate methods are often needed, with additional assumptions such as bounded tree-width~\citep{nie2014advances}. Sampling 
\citep{madigan1995bayesian,grzegorczyk2008improving,niinimaki2012partial,he2016structure} 
and topological order search \citep{friedman2000being,teyssier2012ordering,scanagatta2015learning} are also popular.


In this work we assume that all variables are continuous, and study the DAG structure learning problem with the focus of SEM and smooth loss function $F$ defined over $A$. By using an alternative continuous DAG constraint  $h(A) = 0$, the constrained optimization problem in Eq~\eqref{eqn:original} becomes fully continuous. {An augmented Lagrangian method is then employed in \citep{zheng2018dags,yu2019dag} which solves unconstrained subproblems iteratively to impose the continuous DAG constraint explicitly. 
We investigate whether the explicit DAG constraint can be eliminated entirely, and therefore no iteration would be required.} The key device in accomplishing this is the observation that DAG is associated with curl-free functions on edges $E$ (please see Definition~\ref{def:curl-free} in the supplementary material for a formal definition of curl-free), which motivates a new representation of DAGs with $A=\gamma (W, p)$, $W\in\real^{d\times d}$ and $p\in \real^d$, as will be elaborated in the next section. 
The constrained optimization problem \eqref{eqn:original} can then be replaced by the following objective:
\begin{equation}\label{eqn:2}
(W^*, p^*) = \underset{W, p}{\text{argmin}}\; F( \gamma (W, p), \mathbf{X}), 
\end{equation}
with the optimal DAG $A^*= \gamma (W^*, p^*)$. 
To the best of our knowledge, both the equivalence representation of DAGs and the application of Hodge theory in DAG structure learning have never been studied.


\paragraph{Main Results}
We propose  a new equivalent model for DAGs, discussed in Theorem~\ref{thm:main} and  in Section~\ref{sec:model}, and  show that $\gamma(W,p) = W\circ \relu(\grad(p))$ can be used in Eq~\eqref{eqn:2}, where $\relu$ is the rectified linear unit and $\grad$ is the graph gradient operator.

\begin{thm}\label{thm:main} \textbf{An Equivalent DAG Space.} Consider  a set of $d$ random variables, given any real vector $p\in \real^d$ and any skew-symmetric weight matrix ${W}\in\real^{d\times d}$ satisfying $W=-W^T$, the following (weighted) directed graph spaces are equivalent: 
$${\{\mathcal{G}_{W\circ \relu({\grad(p)})}\}=\mathbb{D} .}$$
\end{thm}

We defer the proof of Theorem \ref{thm:main} to Section \ref{sec:model}, given together by Theorem~\ref{thm:1} and Theorem~\ref{thm:3.7}. To solve \eqref{eqn:2} with the new $\gamma$ function, we further propose a new  algorithm in Section \ref{sec:proj}, based on the combinatorial Hodge theory on graphs \citep{lim2015hodge,jiang2011statistical}. 
As we elaborate details below, we first briefly introduce a few useful basic graph exterior calculus operators 
\citep{bang2008digraphs,jiang2011statistical}, then formulate DAGs into the equivalent model. Due to the page limit, we provide basic definitions for important concepts that are used in this paper in Appendix~\ref{sec:definitions}. For a more thorough introduction of graph calculus, we refer the interested readers to \citep{lim2015hodge}.

Notation-wise, we use $A^0$ to denote the ground truth DAG, $A^*$ for the global optimal solution in Eq~\eqref{eqn:original}, and $\tilde{A}$ for the approximated solution from numerical algorithms. The ultimate goal of our DAG structural learning problem is to obtain a DAG $\tilde{A}$ which recovers the structure of $A^0$ and obtains a comparable score $F(\tilde{A},\mathbf{X})$ as $F(A^*,\mathbf{X})$.

\section{An Equivalent Model for DAGs}\label{sec:model}
 Let $\widehat{\mathcal{G}}=(V,{E})$ be a {complete undirected} 
graph where $V:=\{1,\cdots,d\}$ is the set of vertices and ${E}$ is the set of undirected edges. Note here since $\widehat{\mathcal{G}}$ is a complete graph, the set of $k$-th cliques of $\widehat{\mathcal{G}}$ is equivalent to {all (unordered) subsets of $V$ with size $k$}. 
The ordered and unordered pairs of vertices are delimited by $(i,j)$ and $\{i,j\}$, respectively, where $i$, $j$ denote the $i$-th and $j$-th vertices. Real-valued functions on graphs can be defined on vertices, edges, triangles, and so on. On vertices, a real-valued function $f:V\rightarrow \real$ is called a {potential function}, and we denote the Hilbert space of all potential functions as $L^2(V)$. We may also define real-valued functions on edges $E=\{\{i,j\},i,j\in V\}$ and triangles $T=\{\{i,j,k\},i,j,k\in V\}$, with the requirement that these functions are {alternating}. In particular, an alternating function on edges $Y:V\times V\rightarrow \real$ requires $Y(i,j)=-Y(j,i)$; an alternating function on triangles $\Theta:V\times V\times V\rightarrow \real$, requires $\Theta(i,j,k)=-\Theta(j,i,k)=-\Theta(i,k,j)$. In the following we use $L^2_{\wedge}(E)$ and $L^2_{\wedge}(T)$ to denote the Hilbert spaces of real-valued alternating functions on edges and triangles, respectively. 
Moreover, we note that $p\in L^2(V)$ corresponds to a real vector $p=[p(1),\cdots,p(d)]\in\real^d$, and an alternating function $Y\in L^2_{\wedge}(E)$ corresponds to a skew-symmetric real matrix ${Y}\in\real^{d\times d}$ with ${[Y]}_{ij}=Y(i,j)$ and ${Y}=-{Y}^T$. Here we use the same letter to denote a vector/matrix and the corresponding function on vertices/edges.

Next, we introduce graph calculus operators $\grad$, $\curl$, and their adjoint operators below.
\begin{defn}\label{def:graphop} \cite{lim2015hodge}

The gradient ($\grad: L^2(V)\rightarrow L^2_{\wedge}(E)$) is an operator on any function $p$ on vertices:
$$(\grad\, p)(i,j)=p(j)-p(i), \quad \forall \{i,j\}\in E, $$
and $\grad(p)$ is called a gradient flow. Its adjoint operator ($\grad^*:L^2_{\wedge}(E)\rightarrow L^2(V)$) is defined on any alternating function $Y$ on edges:
$$(\grad^*\, Y)(i)=-\sum_{j=1}^{d}Y(i,j),\quad \forall i\in V.$$
The divergence operator ($\diver:L^2_{\wedge}(E)\rightarrow L^2(V)$) is the negative operator of $\grad^*$, which is also defined  on any alternating function $Y$ on edges:
$$(\diver\, Y)(i)=-(\grad^*\, Y)(i)=\sum_{j=1}^{d}Y(i,j),\quad \forall i\in V.$$
The curl ($\curl: L^2_{\wedge}(E)\rightarrow L^2_{\wedge}(T)$) is an operator for any alternating function $Y$ on edges: 
\begin{align*}
&(\curl\, Y)(i,j,k)=Y(i,j)+Y(j,k)+Y(k,i),\\
&\forall\{i,j,k\}\in T, 
\end{align*}
and its adjoint operator $\curl^*:L^2_{\wedge}(T)\rightarrow L^2_{\wedge}(E)$ is for any alternating function $\Theta$ on triangles:
$$(\curl^*\Theta)(i,j)=\sum_{k=1}^d\Theta(i,j,k), \quad \forall \{i,j\}\in E. $$
The graph Laplacian ($\triangle_0: L^2(V)\rightarrow L^2(V)$) is an operator on any function {$p$} on vertices:
$$(\triangle_0p)(i)=-(\diver\,\grad\,p)(i)=d\cdot p(i)-\sum_{j=1}^{d}p(j), \;\forall i\in V.$$
The graph Helmholtzian ($\triangle_1: L^2_{\wedge}(E)\rightarrow L^2_{\wedge}(E)$) is defined on any alternating function $Y$ on edges:
\begin{align*}
&(\triangle_1Y)(i,j)=(\grad\, \grad^*\,Y+\curl^*\,\curl\,Y)(i,j)\\
&\forall \{i,j\}\in E.
\end{align*}
\end{defn}

\begin{lem}\citep{jiang2011statistical}\label{lemma1}
Let $p\in L^2(V)$ and $\Theta\in L^2_{\wedge}(E)$, denote $D=\grad(p)$ and $R=\curl^*(\Theta)$, then $D$ and $R$ are curl-free and divergence-free, respectively:
\begin{align*}
     &\curl(D)(i,j,k)=0,\quad \forall \{i,j,k\}\in T; \\
&\diver(R) (i)=-\grad^*(R)(i)=0,\quad \forall i\in V.
\end{align*}
\end{lem}
Given a complete undirected graph $\widehat{\mathcal{G}}(V,E)$ and a function $Y\in L^2_{\wedge}(E)$, with $\relu$ denoting the rectified linear unit function, we can define a weighted adjacency matrix $A={\relu(Y)}\in\real^{d\times d}$ as:
\begin{displaymath}
{\relu(Y)}(i,j):=\left\{\begin{array}{cl}
Y(i,j), & \text{ if }\,Y(i,j)>0;\\
0, & \text{ else.}
\end{array}
\right.
\end{displaymath}
and further define a {weighted directed graph} $\mathcal{G}_{\relu(Y)}$ from ${\relu(Y)}$ as the following:
\begin{defn}
Consider a complete undirected graph $\widehat{\mathcal{G}}(V,E)$  and $Y\in L^2_{\wedge}(E)$, {a directed graph $\mathcal{G}_{\relu(Y)}(V,E_{\relu(Y)})$ is defined} such that there is a directed edge from vertex $i$ to vertex $j$ in $\mathcal{G}_{\relu(Y)}$ if and only if $Y(i,j)>0$, i.e., the set of directed edges $E_{\relu(Y)}=\{(i,j)|Y(i,j)>0\}$. Moreover, note that ${\relu(Y)}$ is a weighted adjacency matrix of $\mathcal{G}_{\relu(Y)}$. 
\end{defn}
Based on the above definition, we show that curl-free functions are naturally associated with DAGs:
\begin{lem}\label{lemma2}
Consider a complete undirected graph $\widehat{\mathcal{G}}(V,E)$  and a \textbf{curl-free} function $Y\in L^2_{\wedge}(E)$, then  ${\relu(Y)}\in\real^{d\times d}$ is the weighted adjacency matrix of a DAG. Moreover, given any skew-symmetric matrix ${W}\in\real^{d\times d}$, ${W\circ \relu(Y)}$ is also a DAG, where $\circ$ is the Hadamard product.
\end{lem}
All proofs can be found in the supplemental material. 
Since the gradient of any potential function (gradient flow) is curl-free, with Lemma~\ref{lemma2} the first part of our main theoretical results is obtained:
\begin{thm}\label{thm:1}
Consider $V$ as a set of $d$ random variables, given any real vector $p\in \real^d$ and any skew-symmetric weight matrix ${W}\in\real^{d\times d}$ satisfying $W=-W^T$, ${W\circ \relu({\grad(p)})}$ is the weighted adjacency matrix of a DAG, i.e., {$\{\mathcal{G}_{W\circ \relu({\grad(p)})}\}\subset\mathbb{D}.$}
\end{thm}
Here we note that the proof of Theorem \ref{thm:1} is immediately obtain by taking $Y=\grad(p)$ in  Lemma \ref{lemma2}.

\begin{rmk}
The skew-symmetry requirement of ${W}$ was made based on the fact that at least one or both of $\relu({\grad(p)})(i,j)=0$ and $\relu({\grad(p)})(j,i)=0$ must hold true for any $i,j\in V$. Here we note that $W$ is set to be a skew-symmetric matrix instead of a full matrix so as to have a smaller degrees of freedom ($d(d-1)/2$) in the optimization problem \eqref{eqn:scoreW_main}. In fact, one can use a full matrix (with degrees of freedom $d^2$) or a symmetric matrix (with degrees of freedom $d(d+1)/2$) in place.
\end{rmk}
We now show that the other direction also holds true:
\begin{thm}\label{thm:3.7}
Let ${A}\in\real^{d\times d}$ be the weighted adjacency matrix of a DAG with $d$ nodes, denote $V$ as the corresponding random variables of these $d$ nodes, then there exists a skew-symmetric matrix ${W}\in\real^{d\times d}$, $W=-W^T$, and a real vector $p\in \real^d$ such that ${A}={W\circ \relu({\grad(p)})}$, i.e.,
{$\mathbb{D}\subset\{\mathcal{G}_{W\circ \relu({\grad(p))}}\}.$} {Here $p$ is associated with the topological order of the DAG, such that $p(j)>p(i)$ if there is a directed path from vertex $i$ to $j$.}
\end{thm}


Hence, we have shown that the set of weighted adjacency matrices for DAGs is equivalent to the set of $W\circ \relu({\grad(p)})$, as stated in Theorem~\ref{thm:main}. Denoting $S:=\{W|W \in\real^{d\times d}, W=-W^T\}$ as the space of all $d\times d$ skew-symmetric matrices, 
$\{W\circ \relu({\grad(p)})|W\in S,p\in\real^d\}$ provides an admissible solution set for DAG structural learning problems. 

While we believe our formulation for DAG is general and can have many applications where DAGs are used, in this paper we apply it to the continuous DAG learning framework. With a given loss function $F(A, \mathbf{X})$, the DAG structural learning problem can then be written as an optimization problem without an explicit constraint:
{\begin{equation}\label{eqn:score_main}
    (W^*,p^*)=\underset{W\in S, p\in \real^d}{\text{argmin}} F(W\circ \relu(\grad(p)),\mathbf{X}),
\end{equation}}
and the optimal solution $A^*=W^*\circ \relu(\grad(p^*))$.

The loss function in \eqref{eqn:score_main} is generally nonconvex. Therefore, we inherit the difficulties associated with nonconvex optimization. As will be discussed in the ablation study in Section \ref{sec:lineartest}, numerically solving \eqref{eqn:score_main} with a random initialization of $W$ and $p$ may result in a stationary point which is far from the global optimum. Therefore, instead of solving  \eqref{eqn:score_main} directly, we propose a two-step algorithm, DAG-NoCurl, or NoCurl for short.

\section{Algorithm: DAG with No Curl}\label{sec:proj}

\begin{algorithm}[!t]
 \caption{DAG-NoCurl algorithm}
   \label{alg:proj}
\begin{algorithmic}
\STATE {1:  $A^{pre}\leftarrow$ solve \eqref{eqn:pre}, and threshold $A^{pre}$.}
\STATE {2:  $\tilde{p}\leftarrow$ compute \eqref{eqn:solvep} with $A^{pre}$.}
\STATE {\quad $\tilde{W}\leftarrow$ solve with fixed $\tilde{p}$  \eqref{eqn:scoreW_main}.}
\STATE { \textit{\quad (Full Only)} $\tilde{W}, \tilde{p} \leftarrow$ solve  \eqref{eqn:score_main} with current $\tilde{W}, \tilde{p}$.}
\STATE {\textbf{Return} $\tilde{A} \leftarrow \tilde{W}\circ \relu(\grad(\tilde{p}))$,  and threshold $\tilde{A}$.}
\end{algorithmic}
\end{algorithm}

The proposed DAG-NoCurl has two main steps. In Step 1, we compute an initial estimate of $A^*$, $A^{pre}\in\real^{d\times d}$, whose associated graph $\mathcal{G}_{A^{pre}}$ is not necessarily acyclic. In Step 2, we refine the predicted solution by projecting $A^{pre}$ into the set of DAGs and  obtaining the final solution $\tilde{A}=\tilde{W}\circ \relu({\grad(\tilde{p})})$.  The full algorithm is outlined in Algorithm~\ref{alg:proj}, and  we introduce each step of the algorithm in more details as follows.

\textbf{Step 1:} In order to compute an initial solution  $A^{pre}$, we solve for the following unconstrained optimization problem
{\begin{equation}\label{eqn:pre}
A^{pre} = \underset{A}{\text{argmin}}\;
F(A, \mathbf{X}) + \lambda h(A)
\end{equation}}
where $\lambda>0$ is a constant penalty coefficient, $F(A,\mathbf{X})$ is the loss function and $h(A)$ is a proper continuous constraint for acyclicity. For example, in NOTEARS \citep{zheng2018dags} the authors proposed $h(A)=\tr[\exp(A\circ A)] - d$ and in DAG-GNN \citep{yu2019dag} $h(A)=\tr[(I + \alpha A\circ A)^d] - d$, $\alpha>0$, was employed. Note that when increasing the penalty parameter $\lambda$ to infinity and solving \eqref{eqn:pre} iteratively, an acyclic graph $A$ will be obtained and the procedure will be equivalent to the classical penalty method for constrained optimization problems. However, here  we  solve for \eqref{eqn:pre} with fixed $\lambda$, $h(A^{pre})$ is generally nonzero and $A^{pre}$ is likely not  a DAG. Therefore further computation is required to obtain a DAG. In practice, we find that solving for \eqref{eqn:pre}  at most twice, once with a fixed $\lambda$ ( denoted by the NoCurl-1 cases) or twice with two fixed $\lambda$s ( denoted by the NoCurl-2 cases), gives satisfactory initializations. In the NoCurl-2 cases, we first solve for \eqref{eqn:pre} with a $\lambda=\lambda_1$, and then solve for \eqref{eqn:pre} with a larger fixed penalty coefficient $\lambda=\lambda_2$. 
To achieve a good balance between the computational time and solution accuracy in structural discovery, in Section \ref{sec:lineartest} we perform a hyperparameter study for both one $\lambda$ and two $\lambda$ cases. Note that the DAG constraint in the first step does not need to be strongly enforced; we use the constraint coefficient up to $\lambda=10^3$, which is much smaller than up to $\lambda=10^{16}$  used in NOTEARS.


\textbf{Step 2: } 
{ From Step 1, the learnt $A^{pre}$ is likely to be cyclic. To obtain an acyclic graph solution $\tilde{A}$,  we use the prediction $A^{pre}$ as an initialization and search for its curl-free component by projecting $A^{pre}$ into the admissible solution set $\{{W}\circ\relu(\grad({p}))\}$.}
In particular, we design a projection procedure 
based on the following theorem:
\begin{thm}[Hodge Decomposition Theorem \citep{lim2015hodge,bhatia2012helmholtz,jiang2011statistical}]
Consider an {undirected} graph $\widehat{\mathcal{G}}(V,E)$, any function $Y\in L^2_{\wedge}(E)$ can be decomposed into three orthogonal components:
\begin{equation}\label{eqn:hodge}
Y=\grad(\phi)+\curl^*(\Psi)+H   
\end{equation}
where $\phi \in L^2(V)$, $\Psi\in L^2_{\wedge}(T)$, and $H\in L^2_{\wedge}(E)$. Moreover, $H$ is a harmonic function satisfying $\triangle_1 H=0$, $\curl H=0$ and $\diver H=0$. Here $\triangle_1$ is the graph Helmholtzian operator as defined in Definition \ref{def:graphop}.
\end{thm}

The Hodge decomposition shows that every alternating function on edges $Y\in L^2_{\wedge}(E)$ can be decomposed into two orthogonal components: a gradient flow $\grad(\phi)$, which represents the $L^2$-optimal ordering of the vertices, and a divergence-free component, $\curl^*(\Psi)+H$, which measures the inconsistency of the vertices ordering. Hence we define a {$L^2$ projection operator} $\text{Proj}:L^2_{\wedge}(E)\rightarrow \grad(L^2(V))$ as: 
\begin{equation}\label{eqn:phi}
\text{Proj}(Y)=\grad(\phi),
\end{equation}
such that $\langle \text{Proj}(Y),Y-\text{Proj}(Y)\rangle=0$, where $\langle \cdot, \cdot \rangle $ is the standard $L^2$ inner product on $L^2_{\wedge}(E)$:
$\langle Y,Z\rangle:=\sum_{i,j}Y(i,j)Z(i,j)$.

 We note that $\phi$ is only unique up to a constant potential function, i.e., if $\text{Proj}(Y)=\grad(\phi)$, then $\text{Proj}(Y)=\grad(\phi+C)$ also holds. Adding a constant to $\phi$ will yield the same ordering of vertices and the same gradient flow $\grad(\phi)$. For the sake of well-posedness, we determine a unique solution $\phi$ by fixing its value on the last vertex such that $\phi(d)=0$. Taking the divergence of \eqref{eqn:hodge} yields:
\begin{equation}\label{eqn:proj}
\diver(Y)=\diver\,\grad(\phi)=-\triangle_0 \phi,
\end{equation}
where $\triangle_0$ is the graph Laplacian operator as defined in Definition \ref{def:graphop}.

We obtain the following theorem for the solution of $\phi$:
\begin{thm}\label{thm:solvephi}
Consider an {undirected complete} graph $\widehat{\mathcal{G}}(V,E)$ and $Y\in L^2_{\wedge}(E)$, a solution of \eqref{eqn:phi} is given by 
\begin{equation}\label{eqn:lap0}
\phi=-\triangle_0^\dag \diver(Y)=-\triangle_0^\dag \sum_{j}Y(i,j)
\end{equation}
where $\dag$ indicates the Moore-Penrose pseudo-inverse. Fixing $\phi(d)=0$, the matrix representing the graph Laplacian $\triangle_0$ is given by
\begin{displaymath}
[\triangle_0]_{ij}=\left\{\begin{array}{cl}
d-1,     & \text{ if }i=j \text{ and }i,j\neq d,  \\
-1,     & \text{ if }i\neq j \text{ and }i,j\neq d, \\
0,     & \text{otherwise}.
\end{array}\right.
\end{displaymath}
\end{thm}

Note that a key requirement in the above projection operator is that $Y$ has to be an alternating function on edges, which corresponds to a $d\times d$ skew-symmetric matrix. However, $A^{pre}$ is generally not skew-symmetric. Fortunately, any square matrix $M$ can be written uniquely as a symmetric and a skew-symmetric components:
{\small$M=M_{sym}+M_{skew}$} with {\small$M_{sym}=\frac{1}{2}(M+M^T)$} and {\small$M_{skew}=\frac{1}{2}(M-M^T)$}. Let $C(M)$ denote the connectivity matrix \citep{Jurgbook} of a directed graph $M$ such that $[C(M)]_{ij}=1$ only if a directed path exists from vertex $i$ to vertex $j$. We apply the projection to the skew-symmetric component of the {connectivity matrix} of $A^{pre}$, and we will show this operation preserves the topological order of vertices in a DAG.
\begin{thm}\label{thm:solvep}
Let $A\in\real^{d\times d}$ be the weighted adjacency matrix of a DAG with $d$ nodes, then
\begin{equation}\label{eqn:solvep}
p=-\triangle_0^\dag \diver\left(\frac{1}{2}(C(A)-C(A)^T)\right),
\end{equation}
preserves the topological order in $A$ such that $p(j)>p(i)$ if there is a directed path from vertex $i$ to $j$. Moreover, we have $A=W\circ \relu(\grad(p))$ with the skew-symmetric matrix $W$ defined as
\begin{equation}\label{eqn:defW}
[W]_{ij}=\left\{\begin{array}{ll}
0, \,&\text{if } p(i)=p(j) \text{ or }\\ & \quad A(i,j)=A(j,i)=0;\\
\frac{A(i,j)}{p(j)-p(i)}, &\text{if }A(i,j)\neq 0 \text{ and }A(j,i)=0;\\
\frac{A(j,i)}{p(j)-p(i)}, &\text{if }A(i,j)= 0 \text{ and }A(j,i)\neq0.\\
\end{array}\right.
\end{equation}
\end{thm}

A detailed proof is provided in Appendix \ref{sec:proof}. To help the readers better understand the formulations in Theorems \ref{thm:solvephi} and \ref{thm:solvep}, we also provide concrete examples of the projection procedure for several sample  $A^{pre}$, including both acyclic and non-acyclic ones, in Appendix \ref{sec:eg}. 

Generally, from  Theorem \ref{thm:solvep} we can see that {$\tilde{p}=-\triangle_0^\dag \diver\left(\frac{1}{2}(C(A^{pre})-C(A^{pre})^T)\right)$} provides the topological order for an acyclic approximation of $A^{pre}$. 
Given an adjacency matrix $A$ of a DAG, $p$ can be directly computed from $A$ following formulation \eqref{eqn:solvep}. Therefore, when the learnt $A^{pre}$ from Step 1 is a DAG with the correct topological ordering (even though $A^{pre}$ might be different from the ground truth $A^0$), Theorem \ref{thm:solvep} ensures the learning of an accurate $p$ in Step 2. On the other hand, when $A^{pre}$ is not a DAG but contains some correct ordering information, one can still apply formulation \eqref{eqn:solvep} to learn $p$ which encodes an approximated topological ordering of the vertices. Hence, learning $A^{pre}$ in Step 1 ensures the learning of $p$.

To further refine the solution of $W$, instead of employing \eqref{eqn:defW} directly we optimize $\tilde{W}$ via:
{\begin{equation}\label{eqn:scoreW_main}
    \tilde{W}=\underset{W\in S}{\text{argmin}} \;F(W\circ \relu(\grad(\tilde{p})), \mathbf{X}).
\end{equation}}
To enforce the skew-symmetric property on $W$ we only optimize elements in the upper triangular matrix of $W$, while setting  diagonal elements zero and each lower triangular element as negative of the corresponding upper triangular element, i.e., $W_{ij}=-W_{ji}$. 

In the full version of DAG-NoCurl, given the separately optimized $\tilde{W}$ and $\tilde{p}$, we then jointly optimize both together using Equation~\ref{eqn:score_main}.  As shown in Algorithm~\ref{alg:proj}, our two-step approach does not require any iterative procedure to increase the penalty coefficient $\lambda$. The full version of DAG-NoCurl consists of only 3 (if using only one $\lambda$) or 4 (if using two $\lambda$s) unconstrained optimization problems.

\noindent\textbf{Return a DAG Solution:} we get a DAG solution by $\tilde{A}=\tilde{W}\circ \relu(\grad(\tilde{p}))$.
By the standard practice of thresholding in DAG learning problems (see, e.g., \citet{zheng2018dags}), we employ the same procedure to post-process both $A^{pre}$ after Step 1 and $\tilde{A}$ after Step 2.  The thresholding step in NOTEARS is motivated by rounding numerical solutions to obtain an exact DAG and remove false discoveries, while our threshold step aims to remove false discoveries only since our solution is a DAG already. 
Specifically, we threshold the edge weights of a learned $A$ as follows: given a fixed threshold $\epsilon>0$, set $A(i,j)=0$ if $|A(i,j)|<\epsilon$. In all the numerical tests we use a fixed threshold value $\epsilon=0.3$ as suggested by NOTEARS \citep{zheng2018dags}.




\noindent\textbf{Consistency.} 
 While our DAG-NoCurl algorithm can use any loss function as long as DAGs can be represented by an weighted adjacency matrix, we use the same loss as NOTEARS. 
With the same loss and the equivalence of the DAG space per Theorem~\ref{thm:main},  the global minimizer of the full version of DAG-NoCurl  is the same as that of NOTEARS in Eq~\ref{eqn:original}. 
In practice,  the solution is only guaranteed to be a stationary point due to  the nonconvexity associated with the DAG space, which is shared by all other algorithms within the framework \cite{zheng2018dags}. 

\noindent\textbf{Further Efficiency Improvement.} We observe that jointly optimizing  $W$ and $p$ in Eq~\eqref{eqn:score_main} in Step 2 of Algorithm~\ref{alg:proj} is often not needed in practice, as it may not improve upon the solutions from  Eq~\eqref{eqn:scoreW_main}.  Intuitively, since $p$ indicates the order of all nodes, it means that the topological order $\tilde{p}$ from Step 1 and $\tilde{W}$ from Eq~\eqref{eqn:scoreW_main} given $\tilde{p}$ are similar solution as the ones from Eq~\eqref{eqn:score_main}.  We  show the numerical evidence in the  ablation study and other experiments of Section \ref{sec:lineartest}. Hence we use this more efficient version without Eq~\eqref{eqn:score_main} as the DAG-NoCurl algorithm for experiments. 


\section{Empirical Evaluations}\label{sec:test}

We present empirical evaluations to demonstrate the effectiveness of the proposed DAG continuous representation and learning algorithm. Specifically, we conduct experiments on both linear and nonlinear benchmark synthetic  and real-world datasets, and compare the proposed method DAG-NoCurl against competitive baselines. Here we outline the empirical set-up, with more details including all parameter choices provided in the supplementary material.  The code will be  publicly released at  \url{https://github.com/fishmoon1234/DAG-NoCurl}.

\textbf{Datasets.} We first test our algorithm in linear synthetic datasets. We employ similar experimental setups as existing works \citep{zheng2018dags}. In each experiment, a random graph $\mathcal{G}$ is generated by using the Erd\H{o}s--R\'{e}nyi (ER) model or the scale-free (SF) model with $k$ expected edges (denoted as ER$k$ or SF$k$ cases) and uniformly random edge weights to obtain a ground-truth weighted matrix $A^0$. Given $A^0$, we take $n=1000$ i.i.d. samples of $X$ from the linear SEM $X=(A^0)^TX+Z$, where the noise $Z\in\real^d$ is generated from two different noise models: Gaussian and Gumbel. To apply the NoCurl algorithm in the linear SEM, 
we use the least-square loss  {\small $F(A,\mathbf{X})=\frac{1}{2n} ||\mathbf{X}-A^T\mathbf{X}||_F^2$}
and numerically solve unconstrained optimization problems with L-BFGS \citep{liu1989limited}, although we note that other standard smooth optimization schemes can also be employed. We have also tested loss functions with a smooth $L^1$ regularization in implementation. However results show little improvements. We suspect that using the DAG regularization and a standard threshold procedure may have similar sparsity effect \cite{zheng2018dags} -- a similar observation can be found in regression problems \cite{jain2014}. 

We also test nonlinear SEM and real datasets in Section~\ref{sec:data2} and \ref{sec:data2_real}.  The graphs in the synthetic datasets can be identified exactly in these  SEM settings with additive noises  \cite{peters2014causal} and hence there is no Markov equivalence. 

\textbf{Metrics and Baselines}. We use Structure Hamming Distance (SHD) \citep{zheng2018dags} to show the structure learning accuracy. 
To assess the ability of methods in solving the original optimization problem in Eq~\eqref{eqn:original}, we also report its score difference from the ground truth: $\Delta F=F(\tilde{A},\mathbf{X})-F({A}^0,\mathbf{X})$. For each graph-noise type combination, 100 trials are performed. The exact numerical values for mean accuracy, mean CPU time (in seconds), and mean score difference along with their perspective standard errors of the mean are reported in full in  the supplementary materials. We compare our method with fast greedy equivalent search (FGS) \citep{ramsey2017million}, Causal Additive Models \citep{buhlmann2014cam},  MMPC \citep{tsamardinos2006max}, and NOTEARS \citep{zheng2018dags}. In the supplemental materials, we also compare our method with Equal Variance DAG variants \citep{chen2019causal} , which is designed to handle data with equal variances. For  nonlinear datasets, we also compare with several neural methods and generalized score GES (GSGES) \citep{huang2018generalized}. 


\subsection{Hyperparameter Study}\label{exp:hyper}





We first perform a quantitative study on the hyperparameter $\lambda$ choices. We use the  ER3-Gaussian and ER6-Gaussian cases to select  hyperparameters. We test many sets of hyperparameters and due to the page limit the full numerical results are listed in Table~\ref{Table:hyperER3} for ER3-Gaussian and Table~\ref{Table:hyperER6} for ER6-Gaussian cases in the supplement. It is observed that as long as $\lambda \geq 10$, the accuracy results are all satisfactory. Among which, $\lambda=10^2$ and $\lambda=(10, 10^3)$ are generally the best values in term of both accuracy and computational efficiency. 
Hence, we select them as the default values and refer them as \textbf{NoCurl-1} and \textbf{NoCurl-2}, respectively. For more discussion, please refer to the supplement.


\subsection{Ablation Study} \label{sec:lineartest}

\begin{table}[!t]\renewcommand{\arraystretch}{0.9}
\centering
\caption{ Ablation Study: results (mean $\pm$ standard error over 100 trials) for $d=30$ ER6-Gaussian Cases from DAG-NoCurl, where bold numbers highlight the best method for each case. Lower is better in both time and SHD. Note the usage of color in the table.}
\label{Table:ablation}
\vskip 0.1in
\small
\begin{tabular}{lcccc}
\hline
Method&Time (Sec)&SHD&\\
\hline

rand init&   6.04 $\pm$   0.26&  84.88 $\pm$   2.65\\
rand $p$&  7.74 $\pm$   0.38&  130.37 $\pm$   1.66\\
NoCurl-1{\color{red}s}& \bf{1.58 $\pm$   0.10}& 37.36 $\pm$   1.34 \\
NoCurl-2{\color{red}s}&    4.69 $\pm$   0.23&  29.88 $\pm$   1.55\\
NoCurl-1{\color{red}-}&   1.69 $\pm$   0.11&  32.82 $\pm$   1.07 \\
NoCurl-2{\color{red}-}&  4.67 $\pm$   0.23&  26.08 $\pm$   1.07 \\
NoCurl-1{\color{red}+}& 5.32 $\pm$   0.35&   21.44 $\pm$   1.56\\
NoCurl-2{\color{red}+}&   10.38 $\pm$   0.23&   17.81 $\pm$   1.29\\
\hline
NoCurl-1&    3.34 $\pm$   0.21&  21.44 $\pm$ 1.56 \\
NoCurl-2&  7.68 $\pm$ 0.39&  \bf{ 17.37 $\pm$  1.18}\\
\hline\vspace{-0.15in}
\end{tabular}
\end{table}

We conduct an ablation study on our proposed algorithm, listed in Algorithm~\ref{alg:proj}, testing how each step would affect the final result, with five settings: {1)} solving the optimization problem Eq~\eqref{eqn:score_main} directly with random initialization of $(W,p)$ (denoted as ``\textbf{rand init}''); {2)} NoCurl without Step 1, by solving for $W$ with a random $p$ then performing one additional step to jointly optimize $(W, p)$ with Eq~\eqref{eqn:score_main} (denoted as ``\textbf{rand $p$}''); {3)} NoCurl without Step 2, by repeatedly increasing the threshold of the structure until a DAG is obtained (denoted by \textbf{an additional ``s''} in the end). We use the thresholds starting from $0.3$ (anything below produces much worse results) and with increments of 0.05 until $h(A)<10^{-8}$; {4)} NoCurl with Eq~\eqref{eqn:defW} instead of Eq~\eqref{eqn:scoreW_main} to find $\tilde{W}$ (denoted as  \textbf{an additional ``-''} in the end); {5)} the full version of NoCurl with the step to jointly optimize $(W, p)$ by solving Eq~\eqref{eqn:score_main} after Step 2 (denoted as  \textbf{an additional ``+''} in the end).

We list the results of $d=30$ in ER6-Gaussian in Table~\ref{Table:ablation}, and full numerical results are shown in Table~\ref{Table:hyperER3} to \ref{Table:score4} in the supplement. As one can see from Table~\ref{Table:ablation}, NoCurl with random initializations (``rand init" and ``rand p") performs subpar, indicating the importance of Step 1 in Algorithm~\ref{alg:proj}. Results from NoCurl without Step 2 (NoCurl-1s and NoCurl-2s) are poorer than the full algorithm (NoCurl-1  and NoCurl-2),  indicating that optimizing Eq~\eqref{eqn:scoreW_main} after an initialization is important to refine the solution for better accuracy.  Moreover, results from  NoCurl-1- and NoCurl-2-   also show the important effects of Step 2 of our algorithm: using  Eq~\eqref{eqn:scoreW_main} to learn $\tilde{W}$ instead of Eq~\eqref{eqn:defW} further improves the accuracy. 
As can be seen from {\it$\#$Missing Edge} in Table~\ref{Table:score1} to \ref{Table:score4} in the supplement, Step 2  of our algorithm generally reduces the number of missing edges in comparison to NoCurl-$\cdot$s and NoCurl-$\cdot$-  alternatives. Lastly, adding extra optimization steps (NoCurl-1+ and NoCurl-2+), which are the full version of DAG-NoCurl,  does not result much improvements on accuracy or $\Delta F$. 
This result indicates that 
the efficient version of NoCurl reaches similar solutions in practice. {All discussions above are general and consistent for all dataset tested, as shown in Table~\ref{Table:hyperER3} to \ref{Table:score4} in the supplement.} 

\begin{table*}[!h]\renewcommand{\arraystretch}{0.9}
\begin{center}
\caption{Comparison on score differences (lower is better) from the ground truth, $\triangle F=F(\tilde{A},\mathbf{X})-F({A}^0,\mathbf{X})$.}
\label{Table:miniscore}
\vskip 0.1in
\small
\begin{tabular}{c|cccccc}
\hline
&d=10&d=50&d=100&d=10&d=50&d=100\\
Method&ER3,Gauss &ER3,Gauss &ER3,Gauss&ER6,Gauss &ER6,Gauss &ER6,Gauss\\
\hline
GOBNILP&$-0.03\pm0.00$& - & -&$-0.03\pm0.00$&- &-\\
NOTEARS&$0.03\pm0.01$&$-0.25\pm0.04$&$-1.65\pm0.08$&$0.22\pm0.40$&$1.97\pm0.26$&$2.49\pm0.37$\\
NoCurl-1&$0.09\pm0.02$&$0.05\pm0.05$&$-0.82\pm0.16$&$0.54\pm0.22$&$2.31\pm 0.41$&$4.30\pm0.99$\\
NoCurl-2&$0.06\pm0.02$&$-0.10\pm0.05$&$-1.44\pm0.09$&$0.36\pm0.07$&$1.77\pm0.38$&$2.61\pm0.93$\\
\hline
\end{tabular}
\end{center}
\end{table*}

\subsection{Optimization Objective Results}
\label{exp:lineartest}

\begin{figure*}[!h]
\begin{center}
\subfigure{{\includegraphics[width=0.99\textwidth]{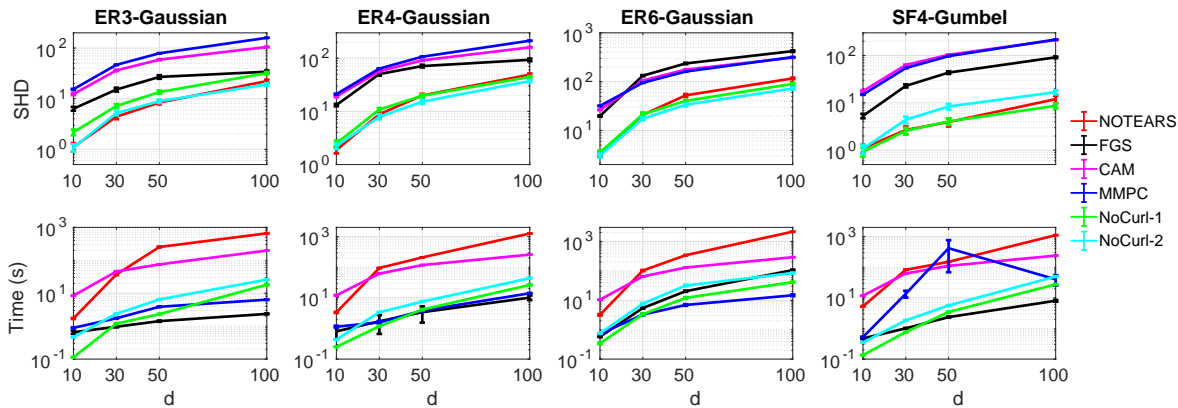}}}
\end{center}
  \caption{Structural discovery  results in terms of SHD (lower is better) and computational time in seconds on linear SEM datasets (log-scale). Error bars represent standard errors over 100 simulations.}
  \label{fig:nocurl_linear}
\end{figure*}

We now study the performance of NoCurl in approximately solving the optimization problem given by \eqref{eqn:original}, by comparing the scores from NoCurl solution $\tilde{A}$ with those from NOTEARS and the exact global minimizer from the GOBNILP algorithm \citep{cussens2016polyhedral}. Since GOBNILP involves enumerating all possible parent sets, its experiments are limited to small DAGs with $d=10$ cases. 
In Table \ref{Table:miniscore}, we show the relative score $\Delta F$ with respect to the ground truth graph $A^0$. Surprisingly, although NoCurl provides an approximate solution, we can see that the score from NoCurl-2 case is very close to the objective values of NOTEARS and GOBNILP, and even outperforms NOTEARS in the ER6 $d=50$ case. When $d$ increases and the optimization problem becomes more difficult,  NoCurl remains competitive and does not deteriorate.

\subsection{Structure Recovery Results}\label{sec:structure}
We now present the comparison with other baselines methods  
by comparing structure recovery accuracy and computational efficiency of NoCurl with NOTEARS, FGS, CAM, and MMPC. 
In Figure \ref{fig:nocurl_linear}, the top row shows the SHD of different methods while the buttom row shows the CPU time, in seconds, of different methods, both in log scales. {For the detailed results in all graph types, please refer to Table \ref{Table:score1} to \ref{Table:score4} in the supplement.}

Consistent with existing observations  \citep{zheng2018dags, buhlmann2014cam}, FGS, MMPC, and CAM's performances suffer when the number of edges gets larger. While NOTEARS is significantly more accurate than other baselines, NoCurl achieves a similar accuracy as NOTEARS, and can sometimes beat NOTEARS, especially on dense and large graphs. For instance,  when $d=100$, in all ER$k$-Gaussian cases the NoCurl-2 achieves the lowest SHD, while in SF4-Gumbel cases the NoCurl-1 achieves the lowest SHD among all methods. When comparing the efficiency, NoCurl requires a similar runtime as FGS and MMPC, which is faster than NOTEARS by more than one or two orders of magnitude. Overall, NoCurl substantially improves the efficiency comparing with NOTEARS, while still sustaining the comparable structure discovery accuracy. 

In addition, we also test  our method with Equal Variance DAG learning algorithm and its variants (EqV-TD and EqV-BU) \citep{chen2019causal}, with their results shown in Table~\ref{Table:score1} to \ref{Table:score4} in the supplement. Our method outperforms both EqV variants by a significant margin. 

\subsection{Nonlinear Synthetic Datasets }\label{sec:data2}

We further test the capability of NoCurl with more general models and datasets by sampling $X$ from nonlinear SEM  $X_j=f(A,X_{pa(j)})+Z_j$ for $j=1,\cdots, d$ with nonlinear functions $f$, following the same setting as \citet{yu2019dag}. For the nonlinear SEM, we combine NoCurl with DAG-GNN, where nonlinear SEM is learnt using neural networks and the standard machinery of augmented Lagrangian was applied to enforce the continuous constraint. We generated 3 Datasets (denoted as Nonlinear Case 1 to 3), and for the exact modeling and implementation details, please refer to the supplement. 

For nonlinear SEM datasets, we  compare NoCurl with DAG-GNN with CAM, MMPC, GSGES, and recent neural-based methods DAG-GNN \citep{yu2019dag}, GraN-DAG \citep{lachapelle2019gradient} and NOTEARS-MLP  \citep{zheng2020learning}. The results are shown in Table~\ref{Table:gnn} in  the supplement. Generally neural-based models outperform heuristic-based methods. Comparing with DAG-GNN, NoCurl has similar accuracy performance across different variable sizes $d$. NoCurl (along with DAG-GNN) is better than NOTEAR-MLP in Nonlinear Case 1 but does not perform as well as GraN-DAG. NoCurl achieves the best overall performance in Nonlinear Case 2. In Nonlinear Case 3, NoCurl is worse than NOTEARS-MLP but much better than GraN-DAG. Regarding the computational time, NoCurl achieves about $3\sim4$ times computational efficiency gain over its base model DAG-GNN on average. NoCurl is also more than one order of magnitude faster than NOTEARS-MLP  and GraN-DAG. Note that NoCurl's performance is limited by the base model, and we choose DAG-GNN as the base model to combine with NoCurl because other neural methods (such as NOTEARS-MLP and GraN-DAG) use a gradient-based adjacency matrix representation, which is different from the weighted $A$ formulation in NoCurl. It would be an interesting future work to extend NoCurl to these frameworks.



\subsection{Real-World Dataset}\label{sec:data2_real}

We now apply NoCurl+DAG-GNN to a real-world bioinformatics dataset~\citep{sachs2005causal} for the discovery of a protein signaling network based on expression levels of proteins and phospholipids. This is a widely used dataset for research on graphical models, with experimental annotations accepted by the biological research community. 
Based on $n=7466$ samples of $d=11$ cell types, 20 edges were estimated in the ground truth graph \citep{sachs2005causal}. In Table \ref{Table:protein} of the supplement, we compare our results and baselines against the ground truth offered \citep{sachs2005causal}. The proposed NoCurl+DAG-GNN obtains an SHD of 16 with 18 estimated edges. The learnt graph is also plotted in supplementary materials. Comparing with the other methods reporting a similar number of nonzero edges, NoCurl has a better performance in terms of SHD. 

\section{Conclusion}
We proposed a novel theoretically-justified continuous representation of DAG structures based on graph exterior calculus operators, and proved that this new formulation can represent the adjacency matrices of DAGs without explicit acyclicity constraints, which is often the most intricate part of the optimization. 
We proposed a new algorithm, which we coin DAG-NoCurl, to approximately solve for the unconstrained optimization problem efficiently.  The key step in this approach is based on the Hodge decomposition theorem, which projects a cyclic graph to the gradient of a potential function and obtains a DAG approximation of the graph. Empirically the proposed DAG-NoCurl achieves comparable accuracy but with substantially better computational efficiency than their counter parts with constraints in all the datasets tested. We believe it is a promising new framework for DAG structure learning where both continuous and discrete optimization approaches can be applied.

Assumptions made in the paper (e.g., SEMs and smooth loss functions) are  widely used and studied in the literature, including the NOTEARS and many related methods. We assume smoothness so we could use gradient-based optimization methods (such as LBFGS) in our proposed algorithm (for ease of comparison with baselines). In fact, the smoothness assumption of loss functions can be further relaxed. For instance, the BFGS method is proved to converge for  continuously differentiable loss functions \cite{li2001global}. Moreover, we note that the proposed framework and algorithm is general, which enables the employment of other optimization methods.
For instance, our framework could handle $L^0$ penalties if one were to use optimization methods such as dynamic programming or equivalent search as suggested by \citet{van2013ell_}. Investigation of the new DAG space in these DAG learning frameworks would be an interesting future work. 

\section*{Acknowledgements}
We thank Dr. Jie Chen and Dr. Changhe Yuan for helpful discussions. We thank to anonymous reviewers who have provide helpful comments. Y. Yu's research is supported in part by the National Science Foundation under award DMS 1753031. Part of this work is also supported by DARPA grant FA8750-17-2-0132.

\bibliography{bib}
\bibliographystyle{plainnat}

\newpage

\appendix

\appendix

\section{Related Concepts} \label{sec:definitions}
In this section, we briefly review a few related mathematical concepts in the paper, which can be found  in \citep{bang2008digraphs,jiang2011statistical}. We use the same notation from the main paper. 

\subsection{Concepts in Linear Algebra}

\begin{defn}[Symmetric and Skew-Symmetric]
A real matrix ${Y}\in\real^{d\times d}$ is called symmetric if and only if $[Y]_{ij}=[Y]_{ji}$ for all $i,j\in\{1,\cdots,n\}$. Similarly, $Y$ is called skew-symmetric if and only if $[Y]_{ij}=-[Y]_{ji}$ for all $i,j\in\{1,\cdots,n\}$.
\end{defn}

\subsection{Concepts in Calculus and Graph Calculus}

\begin{defn}[Hilbert Space]
A Hilbert space is a complete vector space with an inner product defined on the space.
\end{defn}

\begin{defn}[Complete Graph]
A complete graph is a simple undirected graph in which every pair of distinct vertices is connected by a unique edge.
\end{defn}

\begin{defn}[Cliques]
For an undirected graph $\widehat{\mathcal{G}}=(V,{E})$, the set of $k-$th cliques $K_k(\widehat{\mathcal{G}})$ is defined by
$$\{i_1,\cdots,i_k\}\in K_k(\widehat{\mathcal{G}})$$
if and only if all pairs of vertices in $\{i_1,\cdots,i_k\}$ are in $E$. Therefore, when $\widehat{\mathcal{G}}$ is a complete graph, the $k-$th cliques of $\widehat{\mathcal{G}}$ is equivalent to ${\left\{\binom{V}{k}\right\}}$.
\end{defn}

\begin{defn}[Alternating Function]
For an undirected graph $\widehat{\mathcal{G}}=(V,{E})$, an alternating function on $k-$th cliques is: $f:V\times\cdots\times V\rightarrow \real$ satisfying
$$f(i_\sigma(1),\cdots,i_\sigma(k))=sgn(\sigma)f(i_1,\cdots,i_k)$$
for all $\{i_1,\cdots,i_k\}\in K_k$ and $\sigma$ is any permutation on $\{1,\cdots,k\}$. Here $sgn(\sigma)$ denotes the sign of $\sigma$ which is $1$ when the parity of the number of inversions in $(i_\sigma(1),\cdots,i_\sigma(k))$ is even, and $sgn(\sigma)=-1$ if the parity of the number of inversions is odd.
\end{defn}

\begin{defn}[$L^2$ Functions]
For an undirected graph $\widehat{\mathcal{G}}=(V,{E})$, the Hilbert space of all potential functions $f:V\rightarrow\real$ is denoted as $L^2(V)$, with the inner product taken to be the standard inner product: for $f,g\in L^2(V)$,
$$\langle f,g\rangle:=\sum_{i=1}^d f(i)g(i).$$
For the $k-$th cliques, we denote the Hilbert space of all alternating functions on $K_k$ as $L^2_\wedge(K_k)$, with the inner product defined as: for $\Theta,\Phi\in L^2_\wedge(K_k)$,
$$\langle \Theta,\Phi\rangle:=\sum_{\{i_1,\cdots,i_k\}\in K_k}\Theta(i_1,\cdots,i_k)\Phi(i_1,\cdots,i_k).$$
\end{defn}

\begin{defn}[Curl-Free, Divergence-Free] 
\label{def:curl-free}
An edge function $f\in L^2_{\wedge}(E)$ is called curl-free if and only if
$$\curl(f)(i,j,k)=0,\quad \forall \{i,j,k\}\in T,$$
or, equivalently, $f\in ker(\curl)$. Similarly, $f\in L^2_{\wedge}(E)$ is called divergence-free if and only if
$$\diver(f) (i)=-\grad^*(f)(i)=0,\quad \forall i\in V,$$
or, equivalently, $f\in ker(\diver)=ker(\grad^*)$.
\end{defn}

\begin{defn}[Harmonic]
An edge function $f\in L^2_{\wedge}(E)$ is called harmonic if and only if
$$\triangle_1(f)(i,j)=0, \quad \forall \{i,j\}\in E,$$
or, equivalently, $f\in ker(\triangle_1)$.
\end{defn}

\subsection{Concepts in Directed Graphs}

\begin{defn}[Connectivity Matrix]
For a directed graph ${\mathcal{G}}=(V,{E})$ with $d$ vertices, its connectivity matrix $C(\mathcal{G})$ is a $d\times d$ matrix such that $[C(\mathcal{G})]_{ij}=1$ if there exists a directed path from vertex $i$ to vertex $j$, and $[C(\mathcal{G})]_{ij}=0$ otherwise.
\end{defn}

\section{Proof of Lemmas and Theorems}\label{sec:proof}

Here we provide the detailed proof of some critical lemmas and theorems in the main paper.

\subsection{Proof of Lemma 3.4 }

\textbf{Lemma 3.4} {\it Consider a complete undirected graph $\widehat{\mathcal{G}}(V,E)$  and a \textbf{curl-free} function $Y\in L^2_{\wedge}(E)$, then  ${\relu(Y)}\in\real^{d\times d}$ is the weighted adjacency matrix of a DAG. Moreover, given any skew-symmetric matrix ${W}\in\real^{d\times d}$, ${W\circ \relu(Y)}$ is also a DAG, where $\circ$ is the Hadamard product.}

\begin{proof}
We prove the lemma by contradiction. Assuming  that there is a cycle in $\mathcal{G}_{\relu(Y)}$ (the graph with weighted adjacency matrix $\relu(Y)$) on an (ordered) set of nodes $(c_1,c_2,\cdots,c_k,c_1)$ and denoting $c_{k+1}:=c_1$ just for notation simplicity, the curl-free property of $Y$ yields
{$$\sum_{i=1}^{k} Y(c_i,c_{i+1}) = \sum_{i=2}^{k-1} \curl (Y) (c_1,c_{i},c_{i+1})=0.$$ }
There exists at least 1 pair of $(c_i,c_{i+1})$ such that $Y(c_i,c_{i+1})\leq 0$ and hence $(c_i,c_{i+1})\notin E_{\relu(Y)}$, which contradicts with the assumption that $(c_1,c_2,\cdots,c_k,c_1)$ forms a cycle.
\end{proof}
 
\subsection{Proof of Theorem 3.7}

\textbf{Theorem 3.7} {\it Let ${A}\in\real^{d\times d}$ be the weighted adjacency matrix of a DAG with $d$ nodes, denote $\widehat{\mathcal{G}}(V,E)$ as the complete undirected graph on these $d$ nodes, then there exists a skew-symmetric matrix ${W}\in\real^{d\times d}$ and a potential function $p\in L^2(V)$ such that ${A}={W\circ \relu({\grad(p)})}$, i.e.,
$\mathbb{D}\subset\{\mathcal{G}_{W\circ \relu({\grad(p))}}\}.$ Here $p$ is associated with the topological order of the DAG, such that $p(j)>p(i)$ if there is a directed path from vertex $i$ to $j$.}

 \begin{proof}
We first show that there exists a $p\in L^2(V)$ such that
{\begin{equation}\label{eqn:gradfA}
(\grad(p)) (i,j)>0, \; \text{ when }A(i,j)\neq 0.
\end{equation}}
Since $\mathcal{G}_A$ is a DAG, there exists at least one topological (partial) order for its vertices \citep{bang2008digraphs}. 
Taking an topological (partial) order $\mathcal{	\prec}=(c_1,c_2,\cdots,c_d)$ of all the vertices in $\mathcal{G}_A$, $p$ defined as $p(c_i)=i$ satisfies condition \eqref{eqn:gradfA}.
We now construct the weight matrix $W$. Since $A$ represents a DAG, for any two vertices $i$ and $j$, at least one or both of $A(i,j)=0$ and $A(j,i)=0$ must hold true. We define an skew-symmetric matrix $W$ as:
{\begin{equation}\label{eqn:AtoW}
[W]_{ij}=\left\{\begin{array}{l}
0, \,\text{if } p(i)=p(j) \text{ or }A(i,j)=A(j,i)=0;\\
\frac{A(i,j)}{p(j)-p(i)}, \text{if }A(i,j)\neq 0 \text{ and }A(j,i)=0;\\
\frac{A(j,i)}{p(j)-p(i)}, \text{if }A(i,j)= 0 \text{ and }A(j,i)\neq0.\\
\end{array}\right.
\end{equation}}
Then $A=W\circ \relu(\grad(p))$, and we have proved the conclusion.
Moreover, combining Theorem 3.5 and Theorem 3.7, we note that
{$$\mathbb{D}=\{\mathcal{G}_{W\circ \relu({\grad(p))}}\},$$}
which is our main theoretical result.
\end{proof}

\subsection{Proof of Theorem 4.3}

\textbf{Theorem 4.3 }{\it Let $A\in\real^{d\times d}$ be the weighted adjacency matrix of a DAG with $d$ nodes, then
{\begin{equation}\label{eqn:solvep1}
p=-\triangle_0^\dag \diver\left(\frac{1}{2}(C(A)-C(A)^T)\right),
\end{equation}}
preserves the topological order in $A$ such that $p(j)>p(i)$ if there is a directed path from vertex $i$ to $j$. Moreover, we have $A=W\circ \relu(\grad(p))$ with the skew-symmetric matrix $W$ defined as in \eqref{eqn:AtoW}.}

\begin{proof}
Taking any two vertices $i$, $j$ with a directed path from $i$ to $j$, we show that $p(j)>p(i)$. We assume that $i,j\neq d$ without loss of generality, since the proof can be trivially extended to the cases of $i=d$ or $j=d$. Since $C(A)$ is the connectivity matrix of $A$ and $A$ is the weighted adjacency matrix of a DAG, we have the following facts hold:
{\begin{equation}\label{eqn:factC}
[C(A)]_{ii}=[C(A)]_{jj}=[C(A)]_{ji}=0,\, [C(A)]_{ij}=1.
\end{equation}}
Moreover, for any other vertex $k$, if there exists a directed path from $j$ to $k$,  there is also a directed path from $i$ to $k$. Therefore, $[C(A)]_{jk}=1\Rightarrow [C(A)]_{ik}=1$, i.e., $[C(A)]_{ik}\geq [C(A)]_{jk}$. On the other hand, if there exists a directed path from $k$ to $i$, there is also a directed path from $k$ to $j$. Therefore $[C(A)]_{ki}=1\Rightarrow [C(A)]_{kj}=1$ and $[C(A)]_{kj}\geq [C(A)]_{ki}$.

From the definition of $p$ we note that
{$$-\triangle_0 p= \diver\left(\frac{1}{2}(C(A)-C(A)^T)\right).$$}
The $i$-th and $j$-th rows of the above system write:
{\begin{align*}
-dp(i)+\sum_{k=1 }^d p(k)=\frac{1}{2}\left(\sum_{k=1}^d [C(A)]_{ik}-\sum_{k=1}^d [C(A)]_{ki}\right),\\
-dp(j)+\sum_{k=1}^d p(k)=\frac{1}{2}\left(\sum_{k=1}^d [C(A)]_{jk}-\sum_{k=1}^d[C(A)]_{kj}\right).
\end{align*}}
Subtracting the above two equations from each other and applying the facts in \eqref{eqn:factC} yield
{\begin{align*}
d(p(j)-p(i))=&\frac{1}{2}\sum_{k\neq i,j} ([C(A)]_{ik}+[C(A)]_{kj}-[C(A)]_{jk} \\ & -[C(A)]_{ki})+[C(A)]_{ij}-[C(A)]_{ji}\geq 1.
\end{align*}}
Therefore $p(j)>p(i)$, and $A=W\circ \relu(\grad(p))$ can be similarly proved as in Theorem 3.7.
\end{proof}

\section{Complexities}
The computational and space complexities depend on the optimization method used. Let $K$ be the time complexity to solve one objective. For example, for L-BFGS  $K = O(mn)$, where $n$ is the number of variables and $m$ is the number of steps stored in memory. Our method takes $O(K)$ time, compared to $O(LK)$ of NOTEARS where $L$ is the number of iteration in the augmented Lagrangian method. We share the same space complexity as NOTEARS.

\section{Examples of Graph Projection}\label{sec:eg}

\begin{figure}[!h]
\begin{center}
\subfigure{{\includegraphics[width=0.15\textwidth]{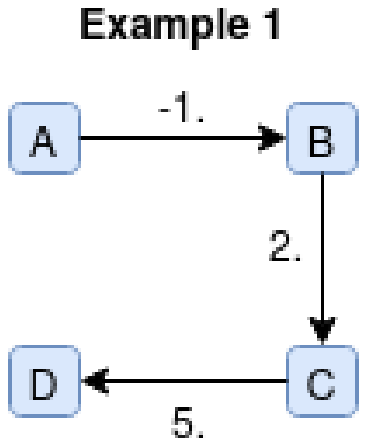}}}\qquad
\subfigure{{\includegraphics[width=0.15\textwidth]{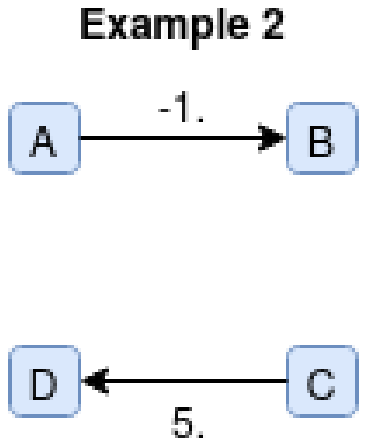}}}\qquad
\subfigure{{\includegraphics[width=0.15\textwidth]{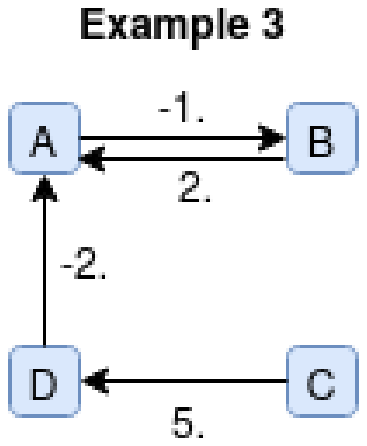}}}\qquad
\subfigure{{\includegraphics[width=0.15\textwidth]{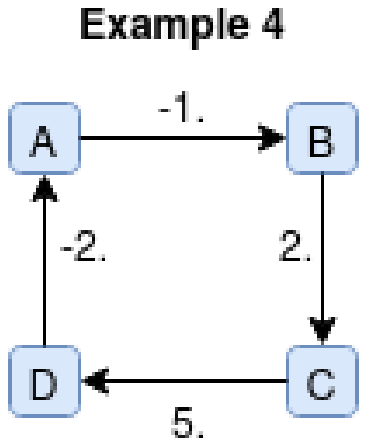}}}
\end{center}
  \caption{Representative graphs as examples to demonstrate the graph projection procedure.}
  \label{fig:eg}
\end{figure}

In this section we provide the detailed procedure of graph projection described in Theorems \ref{thm:solvephi} and \ref{thm:solvep}, for four representative graphs as shown in Figure \ref{fig:eg}. In all examples we consider graphs with $4$ vertices which are denoted as $A$, $B$, $C$ and $D$ in Figure \ref{fig:eg}. In the following calculations we assume $A$ as the first vertex and $D$ as the last vertex. In each example, for a given graph $\mathcal{G}_{A^{pre}}$, we first calculate its approximated gradient flow component via
\begin{equation*}
\tilde{p}=-\triangle_0^\dag \diver\left(\frac{1}{2}(C(A^{pre})-C(A^{pre})^T)\right),
\end{equation*}
then the weights $\tilde{W}$ are computed from \eqref{eqn:defW}. We note that the matrix for graph Laplacian $\triangle_0$ given by \eqref{eqn:lap0} writes:
\begin{displaymath}
[\triangle_0]=\left[\begin{array}{cccc}
3&-1&-1&0\\
-1&-3&-1&0\\
-1&-1&-3&0\\
0&0&0&0\\
\end{array}\right].
\end{displaymath}
Since $\tilde{p}(4)=0$ is fixed, we only need to invert the submatrix of $[\triangle_0]$ formed by ignoring its $4$-th row and $4$-th column, and this submatrix is invertible. Hence the calculation described in Theorem \ref{thm:solvephi} is well-posed.

\textbf{Example 1, projection for a connected acyclic graph (a tree): }We first consider a fully connected acyclic graph as shown in the first plot of Figure \ref{fig:eg}, with the weighted adjacency matrix:
\begin{displaymath}
A^{pre}=\left[\begin{array}{cccc}
0&-1&0&0\\
0&0&2&0\\
0&0&0&5\\
0&0&0&0\\
\end{array}\right].
\end{displaymath}
The connectivity matrix of $A^{pre}$ writes
\begin{displaymath}
 C(A^{pre})=\left[\begin{array}{cccc}
0&1&1&1\\
0&0&1&1\\
0&0&0&1\\
0&0&0&0\\
\end{array}\right].
\end{displaymath}
Therefore, the projection result $\tilde{p}$ from \eqref{eqn:solvep} and the weights $\tilde{W}$ from \eqref{eqn:defW} are obtained:
\begin{displaymath}
 \tilde{p}=\left[\begin{array}{c}
-0.75\\
-0.5\\
-0.25\\
0\\
\end{array}\right],\;\quad\tilde{W}=\left[\begin{array}{cccc}
0&-4&0&0\\
4&0&8&0\\
0&-8&0&20\\
0&0&-20&0\\
\end{array}\right].
\end{displaymath}
We then have the acyclic approximation of $A^{pre}$ as
\begin{displaymath}
 \tilde{A}=\tilde{W}\circ \relu(\grad(\tilde{p}))=\left[\begin{array}{cccc}
0&-1&0&0\\
0&0&2&0\\
0&0&0&5\\
0&0&0&0\\
\end{array}\right]=A^{pre}.
\end{displaymath}
Therefore, the projected potential function $\tilde{p}$ fully preserves the vertices ordering in this acyclic graph, which is consistent with Theorem \ref{thm:solvep}.

\textbf{Example 2, projection for a disconnected acyclic graph (a forest): }Consider an acyclic graph consisting of two trees as shown in the second plot of Figure \ref{fig:eg}, with the weighted adjacency matrix:
\begin{displaymath}
A^{pre}=\left[\begin{array}{cccc}
0&-1&0&0\\
0&0&0&0\\
0&0&0&5\\
0&0&0&0\\
\end{array}\right].
\end{displaymath}
The connectivity matrix of $A^{pre}$ writes
\begin{displaymath}
 C(A^{pre})=\left[\begin{array}{cccc}
0&1&0&0\\
0&0&0&0\\
0&0&0&1\\
0&0&0&0\\
\end{array}\right].
\end{displaymath}
Therefore, the projection result $\tilde{p}$ from \eqref{eqn:solvep} and the weight $\tilde{W}$ from \eqref{eqn:defW} are
\begin{displaymath}
 \tilde{p}=\left[\begin{array}{c}
-0.25\\
0\\
-0.25\\
0\\
\end{array}\right],\;\quad\tilde{W}=\left[\begin{array}{cccc}
0&-4&0&0\\
4&0&0&0\\
0&0&0&20\\
0&0&-20&0\\
\end{array}\right].
\end{displaymath}
We then have the acyclic approximation of $A^{pre}$ as
\begin{displaymath}
 \tilde{A}=\tilde{W}\circ \relu(\grad(\tilde{p}))=\left[\begin{array}{cccc}
0&-1&0&0\\
0&0&0&0\\
0&0&0&5\\
0&0&0&0\\
\end{array}\right]=A^{pre}.
\end{displaymath}
The results indicate that the projected potential function $\tilde{p}$ fully preserves the (partial) ordering in each tree, and the projection procedure in Theorem \ref{thm:solvep} maps the acyclic graph to itself.

\textbf{Example 3, projection for a cyclic graph with cycle length 2: } We now consider a cyclic graph with a  cycle between the first and the second vertex, as shown in the third plot of Figure \ref{fig:eg}, with the weighted adjacency matrix:
\begin{displaymath}
A^{pre}=\left[\begin{array}{cccc}
0&-1&0&0\\
2&0&0&0\\
0&0&0&5\\
-2&0&0&0\\
\end{array}\right].
\end{displaymath}
The connectivity matrix of $A^{pre}$ writes
\begin{displaymath}
 C(A^{pre})=\left[\begin{array}{cccc}
1&1&0&0\\
1&1&0&0\\
1&1&0&1\\
1&1&0&0\\
\end{array}\right].
\end{displaymath}
Therefore, the projection result $\tilde{p}$ and the weight $\tilde{W}$ are
\begin{displaymath}
 \tilde{p}=\left[\begin{array}{c}
0.375\\
0.375\\
-0.25\\
0\\
\end{array}\right],\;\quad \tilde{W}=\left[\begin{array}{cccc}
0&0&0&\frac{16}{3}\\
0&0&0&0\\
0&0&0&20\\
-\frac{16}{3}&0&-20&0\\
\end{array}\right].
\end{displaymath}
We then have the acyclic approximation of $A^{pre}$ as
\begin{displaymath}
 \tilde{A}=\tilde{W}\circ \relu(\grad(\tilde{p}))=\left[\begin{array}{cccc}
0&0&0&0\\
0&0&0&0\\
0&0&0&5\\
-2&0&0&0\\
\end{array}\right]\neq A^{pre}.
\end{displaymath}
It can be seen that when there is a local cycle (between nodes $A$ and $B$ in this example), the projection procedure in Theorem \ref{thm:solvep} simply removes all edges involved in this cycle and keeps the ordering of vertices from all other edges.

\textbf{Example 4, projection for a cyclic graph with cycle length 4: } We now consider a cyclic graph as shown in the last plot of Figure \ref{fig:eg}, with the weighted adjacency matrix:
\begin{displaymath}
A^{pre}=\left[\begin{array}{cccc}
0&-1&0&0\\
0&0&2&0\\
0&0&0&5\\
-2&0&0&0\\
\end{array}\right].
\end{displaymath}
The connectivity matrix of $A^{pre}$ writes
\begin{displaymath}
 C(A^{pre})=\left[\begin{array}{cccc}
1&1&1&1\\
1&1&1&1\\
1&1&1&1\\
1&1&1&1\\
\end{array}\right].
\end{displaymath}
Therefore, $C(A^{pre})-C(A^{pre})^T)=0$, and the projection results $\tilde{p}=(0,0,0,0)^T$. We then have the acyclic approximation of $A^{pre}$ as
\begin{displaymath}
 \tilde{W}\circ \relu(\grad(\tilde{p}))=\left[\begin{array}{cccc}
0&0&0&0\\
0&0&0&0\\
0&0&0&0\\
0&0&0&0\\
\end{array}\right]
\end{displaymath}
This example illustrates that when there is a cycle with length greater than 2, the projection procedure in Theorem \ref{thm:solvep} removes all edges between any two nodes in this cycle.

\section{Detailed Algorithm and Experiment Settings}\label{sec:setting}

\subsection{Settings on Synthetic Dataset}

In the following we briefly describe the empirical process of generating synthetic datasets. The code will be  publicly released at  \url{https://github.com/fishmoon1234/DAG-NoCurl}.

\textbf{Linear synthetic datasets}: In the linear SEM tests, for each $d\in\{10,30,50,100\}$ and each graph type-noise type combination, $100$ trials were performed with $1000$ samples in each dataset. For each trial, a ground truth DAG $\mathcal{G}_{A^0}$ is randomly sampled following either the Erd\H{o}s--R\'{e}nyi (ER) or the scale-free (SF) scheme. When $(i,j)$ is a directed edge of the ground truth DAG $\mathcal{G}_{A^0}$, the weight of this edge $A^0_{ij}$ is sampled from $\mathcal{U}([-2,-0.5]\cup[0.5,2])$. Each sample $X^i\in\real^d$, $i=1,\cdots,1000$, is generated following:
$$X^i_j= (a^0_j)^T \pi^0(X^i_j)+Z_j^i$$
where $X^i_j$ is the $i$th sample of  $j$th variable $X_j$, $a^0_j\in\real^d$ is the $j$th column of the ground truth weighted adjacency matrix $A^0=[a^0_1|\cdots|a^0_d]$, $\pi^0(X_j^i)$ is a random vector of size $d$
containing the variable values corresponding to the parents of  $j$th variable $X_j$ per $A^0$ in  the $i$th sample, i.e., its $k$-th component $[\pi^0(X_j^i)]_k = X^i_k$ if $X_k$ is a parent of $X_j$ in $A^0$ otherwise $[\pi^0(X_j^i)]_k = 0$,
$Z^i_j$ is either a Gaussian noise $Z^i_j\sim \mathcal{N}(0,1)$ or a Gumbel noise $Z^i_j\sim\text{Gumbel}(0,1)$.

\textbf{Nonlinear synthetic datasets:} In the nonlinear SEM tests, $5$ trials were performed for each case with $5000$ samples in each dataset. For each trial the ground truth DAG $\mathcal{G}_{A^0}$ and the weighted adjacency matrix $A^0$ are generated following the same way as in the linear SEM tests. Three types of datasets were considered:
\begin{itemize}
    \item \textit{Nonlinear Case 1}: For each $d\in\{10,20,50,100\}$, each sample $X^i\in\real^d$, $i=1,\cdots,5000$, is generated following 
    $$X^i_j= \cos(  (a^0_j)^T  \pi^0(X^i_j)+1)+Z_j^i$$

    where  $A^0$ is the weighted adjacency matrix of a graph sampled following the Erd\H{o}s--R\'{e}nyi (ER) scheme with $3d$ expected edges (denoted as \textbf{ER3}), and the noise $Z^i_j\sim\mathcal{N}(0,1)$.
    
    \item \textit{Nonlinear Case 2}: For each $d\in\{10,20,50,100\}$, each sample $X^i\in\real^d$, $i=1,\cdots,5000$, is generated following 
    \begin{align*}
    X^i_j=&2\sin((a^0_j)^T\pi^0(X^i_j)+0.5) \\
    &+((a^0_j)^T\pi^0(X^i_j)+0.5)+Z^i_j
    \end{align*}
    where  $A^0$ is the weighted adjacency matrix of a graph sampled following the Erd\H{o}s--R\'{e}nyi (ER) scheme with $3d$ expected edges (denoted as \textbf{ER3}), and the noise $Z^i_j\sim\mathcal{N}(0,1)$.
    
    \item \textit{Nonlinear Case 3}: For each $d\in\{10,20,50,100\}$, each sample $X^i\in\real^d$, $i=1,\cdots,5000$, is generated following 
    $$X^i_j=(a^0_j)^T\cos(\pi^0(X^i_j)+\mathbf{1})+Z^i_j$$
    where  $A^0$ is the weighted adjacency matrix of a graph sampled following the Scale-free (SF) scheme with $3d$ expected edges (denoted as \textbf{SF3}), and the noise $Z^i_j\sim\mathcal{N}(0,1)$.
\end{itemize}
Here we note that Nonlinear Case 1 and 2 were adopted from \citep{yu2019dag}. In Nonlinear Case 3, each sample were generated following almost the same scheme as in Nonlinear Case 1, but the ground truth graph was generated with the SF model. Comparing with the ER graphs which have a degree distribution following a Poisson distribution, SF graphs have a degree distribution following a power law and therefore few nodes have a high degree \citep{lachapelle2019gradient}.

\subsection{Settings for Each Algorithm}

In this section we describe the settings and parameters employed in each algorithm.

\subsubsection{Linear SEM}

\textbf{DAG-NoCurl:} In linear SEM we use the least-squares loss
\begin{equation}\label{eqn:FSEM}
F_{SEM}(A,\mathbf{X})=\frac{1}{2n} ||\mathbf{X}-A^T\mathbf{X}||_F^2
\end{equation}
regardless of the noise type, with the polynomial acyclicity penalty from \citep{yu2019dag}
\begin{equation}\label{eqn:h}
h(A)=\text{tr}[(I+A\circ A/d)^d]-d.
\end{equation}
We consider the penalty parameter $\lambda$ in DAG-NoCurl as a tunable hyperparameter, with the range  of $\{1, 10, 10^2, 10^3, 10^4\}$. We use the runtime and the score difference from the ground truth $\Delta F=F_{SEM}(A,\mathbf{X})-F_{SEM}(A^0,\mathbf{X})$ as the measure to choose the best hyperparameters. For the detailed analysis and discussion, please refer to the  Section \ref{sec:parameter} on hyperparameter study of this supplemental material. To solve for the unconstrained smooth minimization problems, although a number of efficient numerical algorithms are available, we employ the L-BFGS \citep{liu1989limited} algorithm with the stopping tolerance ``ftol'' (the relative score difference between the last two iterations) set as $10^{-8}$. The implementation is in Python based on the original NOTEARS package from \citep{zheng2018dags}. Unless otherwise stated, we use the threshold $0.3$ on $A^{pre}$ and $\tilde{A}$, as suggested in \citep{zheng2018dags}.

\textbf{NOTEARS: }For baseline method NOTEARS, we use the NOTEARS package in Python from \citep{zheng2018dags} with the least-squares loss \eqref{eqn:FSEM} and the polynomial acyclicity penalty \eqref{eqn:h}. For the augmented Lagrangian method in NOTEARS, we use default parameters from the package, and the default stopping criteria $h(A)\leq 10^{-8}$.

\textbf{GOBNILP: } For the exact minimizer of the original optimization problem, we use the publicly available package Globally Optimal Bayesian Network learning using Integer Linear Programming (GOBNILP) \citep{cussens2016polyhedral}\footnote{\url{https://www.cs.york.ac.uk/aig/sw/gobnilp/}}. It uses integer linear programming written in C program and SCIP optimization solvers  to learn BN from complete discrete data or from local scores. We use GaussianL0 score with $k=0.0$ and did not set a maximal parental set size (``palim = None"). We did not change any other   parameter setting.

\textbf{FGS: }For baseline method fast greedy equivalent search (FGS), we use py-causal package from Carnegie Mellon University \citep{ramsey2017million}\footnote{\url{https://github.com/
bd2kccd/py-causal}}. This method is written in highly optimized Java code with a Python interface. We use the default parameter settings and did not tune any parameter. Instead of returning a DAG, a CPDAG is returned by FGS which contains undirected edges. Therefore, in our evaluations for FGS, we favorably treat undirected edges from FGS as true positives, as long as the ground truth graph has a directed edge in place of the undirected edge.

\textbf{CAM: }For baseline method causal additive models (CAM) \citep{buhlmann2014cam}, we use Causal Discovery toolbox in Python\footnote{\url{https://github.com/FenTechSolutions/CausalDiscoveryToolbox}}. Only two input parameters, ``variablesel'' and “pruning", were tuned, which enables preliminary neighborhood selection and pruning, respectively. We found that with the preliminary neighborhood selection applied the time consumption of CAM is reduced significantly, and the pruning step helps reducing the resultant SHD and therefore improves the accuracy. These observations are consistent with the experiments reported in \citep{buhlmann2014cam}. Therefore, all results reported here are with these two parameters turned on.

\textbf{MMPC: }For baseline method Max-Min Parents and Children (MMPC) \citep{tsamardinos2006max}, we also use Causal Discovery toolbox in Python\footnote{\url{https://github.com/FenTechSolutions/CausalDiscoveryToolbox}}, with the default parameter settings.

\textbf{Eq-TD \& Eq-BU:} we use the available code form github\footnote{\url{https://github.com/WY-Chen/EqVarDAG}} and the same named functions as listed. We did not tune any hyperparameters.

\subsubsection{Nonlinear SEM}\label{sec:nonlinearsetting}

\begin{algorithm}[!t]
 \caption{NoCurl algorithm combining with DAG-GNN}
   \label{alg:gnnproj}
\begin{algorithmic}
\STATE {1. Step 1:  Solve for an initial prediction  $(A^{pre},\theta^{pre})$ with
\begin{align*}
(A^{pre},\theta^{pre}) = & \underset{A,\theta}{\text{argmin}}
\,\{-L_{\text{ELBO}}(A,\mathbf{X})\\
&+\lambda(\text{tr}[(I+A\circ A/d)^d]-d)\}
\end{align*}
and threshold $A^{pre}$.}
\STATE {2. Step 2: Based on $A^{pre}$, obtain an approximate solution of $p^*$ as $\tilde{p}$ with 
$$\tilde{p}=-\triangle_0^\dag \diver\left(\frac{1}{2}(C(A^{pre})-C(A^{pre})^T)\right),$$
then solve for $\tilde{W}$ with fixed $\tilde{p}$ via
$$(\tilde{W},\tilde{\theta}) =  \underset{W,\theta}{\text{argmin}}
\,-L_{\text{ELBO}}(W\circ\relu(\grad(\tilde{p})),\mathbf{X})$$
In this step, the initial prediction for parameters $\theta^{pre}$ from Step 1 is used as the initial guess of $\theta$.}
\STATE {3. Obtain the final approximation solution $\tilde{A}=\tilde{W}\circ \relu(\grad(\tilde{p}))$  and threshold $\tilde{A}$.}
 \end{algorithmic}
\end{algorithm}

\textbf{DAG-GNN with NoCurl: }In nonlinear SEM we combine NoCurl with DAG-GNN \citep{yu2019dag}. In DAG-GNN, a deep generative model is employed to learn the DAG by maximizing the evidence lower bound (ELBO):

{\begin{align}
\nonumber &L_{\text{ELBO}}(A,\mathbf{X}) =\frac{1}{n}\overset{n}{\underset{k=1}\sum} L_{\text{ELBO}}^k(A,X^k) \\
\nonumber \text{where }&L_{\text{ELBO}}^k(A,X^k)=-D_{\text{KL}}\Big(q(Y|X^k;A) \,||\, p(Y)\Big)  \\
\nonumber&\qquad\qquad\qquad+\mathbb{E}_{q(Y|X^k;A)}\Big[\log p(X^k|Y;A)\Big].\label{eqn:elbo.i}
\end{align}}

Following the settings in \citep{yu2019dag}, $Y\in\real^d$ is a latent variable and $p(Y)$ is the prior modeled with the standard multivariate normal $p(Y)=\mathcal{N}(0,I)$. $q(Y|X;A)$ is the variational posterior to approximate the actual posterior $p(Y|X)$, and $D_{\text{KL}}$ denotes the  KL-divergence between the variational posterior and the actual one. $q(Y|X;A)$ is modeled with a factored Gaussian with mean $M_Y\in \real^d$ and standard deviation $S_Y\in\real^d$, based on a multilayer perception (MLP):
$$[M_Y|\log S_Y]=(I-A^T)\text{MLP}(X,M^1,M^2)$$
where $M^1\in \real^{1\times n_{hid}}$ and $M^2\in \real^{n_{hid}\times 1}$ are parameters and $n_{hid}$ is the number of neurons in the hidden layer. Similarly, $p(X|Y;A)$ is also modeled with a factored Gaussian with mean $M_X\in \real^d$ and standard deviation $S_X\in\real^d$, based on a multilayer perception (MLP):
$$[M_X|\log S_X]=\text{MLP}((I-A^T)^{-1}Y,M^3,M^4)$$
where $M^3\in \real^{1\times n_{hid}}$ and $M^4\in \real^{n_{hid}\times 1}$ are parameters. In DAG-GNN, the weighted adjacency matrix $A$ is optimized together with all the parameters $\theta=(M^1,M^2,M^3,M^4)$ with the following learning problem:
\begin{equation}\label{eqn:originalGNN}
{
\begin{aligned}
&&(A^*,\theta^*) =  \underset{A,\theta}{\text{argmin}}
\,-L_{\text{ELBO}}(A,\mathbf{X}),\\
&&\text{ subject to }
 h(A)=\text{tr}[(I+A\circ A/d)^d]-d=0. 
\end{aligned}}
\end{equation}

With the goal of boosting the efficiency of DAG-GNN without losing accuracy, in NoCurl we use the same score function from DAG-GNN: {$$F_{\text{ELBO}}(A,\mathbf{X})=-L_{\text{ELBO}}(A,\mathbf{X})$$}
and their implementation based on PyTorch~\citep{paszke2017automatic}. The detailed steps are described in Algorithm \ref{alg:gnnproj}. Specifically, we use DAG-GNN's default number of neurons in the hidden layer $n_{hid}=64$. Each unconstrained optimization problem is solved using the Adam~\citep{Kingma2015}, with the default learning rate$=3e-3$ from DAG-GNN. To guarantee sufficient updates for the parameters $\theta$, we use epoch number$=400$ in Step 1 and epoch number$=600$ while solving for $\tilde{W}$ in Step 2. Due to the computation load of neural models, we use one fixed $\lambda=10$ as the hyperparameter in NoCurl since it is the fastest method while being reasonably accurate per our hyperparameter study on linear SEM datasets (see the hyperparameter study in  Section \ref{sec:parameter} of this supplementary material).

\textbf{DAG-GNN:} We use the available code from github \footnote{\url{https://github.com/fishmoon1234/DAG-GNN}} to run DAG-GNN. We use the default  hidden size  $64$ for all layers and did not tune any other hyperparameters (all default values). 

\textbf{GraN-DAG:} We use the available code from github \footnote{\url{https://github.com/kurowasan/GraN-DAG}}. Following the suggestion in \citep{lachapelle2019gradient}, we turned on both the preliminary neighborhood selection (\texttt{pns}) and the pruning option (\texttt{cam-pruning}), for which we have observed a big improvement in SHD. For the rest of hyperparameters, we use default values with options \texttt{pns} and \texttt{cam-pruning}.

\textbf{NOTEARS-MLP:} We use the available code from github \footnote{\url{https://github.com/xunzheng/notears}} to run NOTEARS-MLP. We tune the hidden size to 32 for all layers (increased from default size 10, for which we observe a big improvement in SHD) to improve the accuracy. All other hyperparameters are kept as their default values: the augmented Lagrangian method terminates when $h(A)=\text{tr}[(I+A\circ A/d)^d]-d\leq 10^{-8}$ and  $\lambda1$ and $\lambda2$ are set as $0.01$. 

For \textbf{CAM} and \textbf{MMPC}, we use the same settings as discussed in the previous subsection. 

\textbf{GSGES:} We use the available code from github \footnote{\url{https://github.com/Biwei-Huang/Generalized-Score-Functions-for-Causal-Discovery}}. We use the default settings for evaluations. 

\subsection{Other Experiment Details}

\textbf{A clear definition of the specific measure or statistics used to report results}: To evaluate the accuracy of results from each algorithm, we mainly use the structure hamming distance (SHD) as a metric, which is the sum of extra, missing, and reverse edges in learned graphs. We report the computational time (in seconds) of each algorithm, as a main metric of their computational efficiency. When it is available, we also report the score difference from the ground truth (denoted as $\Delta F$), the number of extra edges (denoted as \#Extra E),  the number of missing edges (denoted as \#Missing E) and the number of reverse edges (denoted as \#Reverse E). All metrics are the lower the better.

\textbf{A description of results with central tendency (e.g. mean) \& variation (e.g. error bars)}: We report mean and standard error of the mean for each metric, with a format as ``mean $\pm$ standard error''.

\textbf{The average runtime for each result, or estimated energy cost}: We use CPU and report the run time (in seconds) for each algorithm. We run all the algorithms up to 72 hours for each trial.

\textbf{A description of the computing infrastructure used}: We use a local Linux-based computing cluster, and all the codes are written in Python and/or PyTorch.

\section{Hyperparameter Study}\label{sec:parameter}

In this section we continue the discussion on hyperparameter study results in Section \ref{exp:hyper}  of the main text and conduct a hyperparameter study for linear SEMs, with one fixed $\lambda$ or two fixed $\lambda$'s in Step 1 of the proposed algorithm. In particular, in the one fixed $\lambda$ cases (denoted as the $\lambda=\cdot$ cases), we obtain the estimate $A^{pre}$ in Step 1 by solving for only one unconstrained optimization problem:
$$A^{pre} = \underset{A}{\text{argmin}}\;
F(A, \mathbf{X}) + \lambda h(A),$$
where $A\in\real^{d\times d}$ is initialized as $A_{ij}=0$, $\forall i,j\in\{1,\cdots,d\}$. 
In the two fixed $\lambda$'s cases (denoted as the $\lambda=(\lambda_1,\lambda_2)$ cases), we obtain the estimate $A^{pre}$ in Step 1 by solving for two optimization problems sequentially. We firstly solve:
$$A^{pre,0} = \underset{A}{\text{argmin}}\;
F(A, \mathbf{X}) + \lambda_1 h(A),$$
with initial guess $A_{ij}=0$, $\forall i,j\in\{1,\cdots,d\}$, then use $A^{pre,0}$ as the initial guess to solve 
$$A^{pre} = \underset{A}{\text{argmin}}\;
F(A, \mathbf{X}) + \lambda_2 h(A)$$
for the estimate matrix  $A^{pre}$. Here we explore the hyperparameter $\lambda$ on ER3-Gaussian and ER6-Gaussian cases, to investigate the performances of NoCurl on both relatively sparse graphs (ER3) and relatively dense graphs (ER6). The results for ER3-Gaussian are provided in Table \ref{Table:hyperER3} and the results for ER6-Gaussian is in Table \ref{Table:hyperER6}. For all cases we report the structure hamming distance (SHD), the score difference from the ground truth (denoted as $\Delta F$), the run time (in seconds), the number of extra edges (denoted as \#Extra E),  the number of missing edges (denoted as \#Missing E) and the number of reverse edges (denoted as \#Reverse E), while we choose the hyperparameter mainly based on the considerations of both a low run time and a good resultant score from the predicted graph (low $\Delta F$).

For cases with one fixed $\lambda$, we investigate the hyperparameter $\lambda\in[10^0,10^4]$. From Tables \ref{Table:hyperER3} and \ref{Table:hyperER6} it can be observed that in both ER3-Gaussian and ER6-Gaussian cases, comparing with the other values of $\lambda$'s, tests with $\lambda=10$ and $\lambda=10^2$ generally require short run time and their predicted graphs have  relatively good scores according to their resultant loss values $\Delta F$. $\lambda=10$ is faster and more accurate in SHD in ER3 than $\lambda=10^2$ for all $d$, but $\lambda=10^2$ has better (i.e., lower) $\Delta F$ loss values. In denser graphs (ER6), $\lambda=10^2$ becomes significantly better in both $\Delta F$ and SHD. As a result, we use $\lambda=10^2$ as the default hyperparameter value for one fixed $\lambda$, which becomes \textbf{NoCurl-1}, experiments in the following. 

For cases with with two fixed $\lambda$'s, we test the cases with $\lambda_1,\lambda_2\in[10^0,10^4]$, and list some combinations with results in Tables \ref{Table:hyperER3} and \ref{Table:hyperER6}. Specifically, in most cases $\lambda=(10, 10^3)$  and  $\lambda=(10, 10^4)$ are the two combinations with the best score values $\Delta F$. Among these two combinations, we found that $\lambda=(10, 10^4)$ results in slightly lower $\Delta F$ loss and SHD, but $\lambda=(10, 10^3)$ requires a lower run time, especially when $d$ is large. Here we choose $\lambda=(10, 10^3)$ as the default parameters in two fixed  $\lambda$'s experiments, which becomes \textbf{NoCurl-2}. 

{From Tables \ref{Table:hyperER3} and \ref{Table:hyperER6}, we also observe that, to achieve the optimal loss and accuracy, larger and denser graphs generally require a larger value of penalty parameter $\lambda$. As a future direction, we are investigating the strategy of choosing $\lambda$ automatically.}

\section{Ablation Study}\label{sec:ablation}

In this section we continue the discussion on ablation study results in Section \ref{sec:lineartest}  of the main text and perform an ablation study, to investigate the effects of each step in our proposed algorithm. In particular, results from the following five settings are listed in Tables \ref{Table:hyperER3} and \ref{Table:hyperER6}:
\begin{itemize}
    \item \textbf{rand init cases}: We solve for $(\tilde{W},\tilde{p})$ from the optimization problem
    \begin{equation}\label{eqn:score}
    (\tilde{W},\tilde{p})=\underset{W\in S, p\in \real^d}{\text{argmin}} F(W\circ \relu(\grad(p)),\mathbf{X})
    \end{equation}
    directly, with random initialization of $(W,p)$. The results are the average from $7$ different random initializations $W_{ij}\sim \mathcal{U}([0,1])$, $p_i\sim \mathcal{U}([0,1])$ for each set of data. With this test we aim to investigate the importance of both Step 1 and Step 2.
    \item \textbf{rand $p$ cases}: We omit Step 1 and initialize $p^{init}$ with random initializations, then solve for an estimate of $W$ from
    $${W}^{pre}=\underset{W\in S}{\text{argmin}} \;F(W\circ \relu(\grad({p}^{init})), \mathbf{X})$$
    and finally jointly optimize $(\tilde{W},\tilde{p})$ from the optimization problem in \eqref{eqn:score}. The results are also the average from $7$ different random initializations of $p$ following $p_i\sim \mathcal{U}([0,1])$ for each set of data. With this test we aim to investigate the importance of Step 1.
    
    \item \textbf{  $\lambda=10^2$s and $\lambda=(10,10^3)$s cases, which are  NoCurl-1s and Nocurl-2s with ``s"}: We test if Step 2 of the algorithm is important. In particular, we solve Step 1 and then use an incremental thresholding method to obtain a DAG from the potential cyclic graph $A^{pre}$ of Step 1.  In these cases, we repeatedly increase the threshold of the structure until a DAG is obtained. 
    We use the thresholds starting from $0.3$ (anything below produces much worse results) and with increments of $0.05$ until $h(A)<10^{-8}$.
    
    \item \textbf{$\lambda=10^2-$ and $\lambda=(10,10^3)-$ cases,  which are  NoCurl-1- and Nocurl-2- with ``-"}: Instead of solving for $\tilde{W}$ from the optimization problem 
    \begin{equation}\label{eqn:scoreW}
    \tilde{W}=\underset{W\in S}{\text{argmin}} \;F(W\circ \relu(\grad(\tilde{p})), \mathbf{X}),
    \end{equation}
    we estimate $W$ directly from $A^{pre}$ with the formulation \eqref{eqn:AtoW} above. When $A^{pre}$ is a DAG, the formulation \eqref{eqn:AtoW} will fully recover $A^{pre}$. Otherwise, when there is a cycle in $\mathcal{G}_{A^{pre}}$, this formulation will remove all edges between any two nodes in this cycle. With this study we aim to check the importance of the second part of Step 2, i.e., solving for $\tilde{W}$ from \eqref{eqn:scoreW}.
    
    \item \textbf{$\lambda=10^2+$ and $\lambda=(10,10^3)+$ cases, which are NoCurl-1+ and Nocurl-2+ with ``+"}: After Step 1 and Step 2 of our algorithm, We add one additional post-processing step to jointly optimize $(\tilde{W},\tilde{p})$ from the optimization problem in \eqref{eqn:score}, so as to guarantee that the solution is a stationary point of the optimization problem \eqref{eqn:score}. 
    This study aims to investigate how far our approximated solution is from a stationary point.
    
\end{itemize}

As one can see from Tables \ref{Table:hyperER3} and \ref{Table:hyperER6}, NoCurl with random initializations (``rand init") performs subpar, indicating the importance of Step 1  of our algorithm. Among the two random initialization cases, the ``rand $p$'' cases have a even worse accuracy, especially on the number of reserved edges, which indicates that a good estimate of the topological ordering in $p$ plays a critical role in the algorithm. Results from threshold $s$ cases show that they are not as good as the full algorithm, indicating that Step 2 is also critical to the performance of our method. Moreover, we list all threshold $s$ cases from other empirical settings in Table \ref{Table:score1} to \ref{Table:score4} of Section \ref{sec:SHD}, to show that poor results are consistent across different settings. In addition, by comparing the $\lambda=10^2$ case with $\lambda=10^2-$ case and the $\lambda=(10, 10^3)$ case with $\lambda=(10, 10^3)-$ case, we found that although the $\lambda=10^2-$ and $\lambda=(10, 10^3)-$ cases are less likely to predict a wrong extra edge, but their predicted graphs tend to miss a relatively large number of edges and therefore have a large SHD. 
When there is a cycle in $\mathcal{G}_{A^{pre}}$, the formulation \eqref{eqn:AtoW} will remove all edges between any two nodes in this cycle. On the other hand, the numbers of missing edges from  the $\lambda=10^2$ and $\lambda=(10, 10^3)$ cases are much lower, which indicates that the algorithm has successfully recovered some of the lost edges when solving for $\tilde{W}$ from \eqref{eqn:scoreW}. Lastly, by comparing the $\lambda=10^2$ cases with $\lambda=10^2+$ cases and the $\lambda=(10, 10^3)$ cases  with $\lambda=(10, 10^3)+$ cases, we observe that adding extra optimization steps after Step 2 does not result much improvements on accuracy or $\Delta F$. This result indicates that the estimated solution $(\tilde{W},\tilde{p})$ from our algorithm  is often very close to a stationary point of \eqref{eqn:score}.

\section{Optimization Objective Results}\label{sec:opt}

In this section we continue the discussion on optimization objective results in Section \ref{exp:lineartest}  of the main text, by displaying the additional results for optimization objective results $\Delta F=F(\tilde{A},\mathbf{X})-F(A^0,\mathbf{X})$ for different graph-type and noise-type combinations in Table \ref{Table:fullscore}. As one may see, the two fixed $\lambda$ case can achieve close objective values to NOTEARS, while in the denser graph case (ER6) the $\lambda=(10,10^3)$ case even outperforms NOTEARS when $d=30$ and $d=50$. This result is encouraging but also surprising since the problem is often more difficult as the graph becomes larger and denser, and our algorithm only provides an approximated solution. We suspect one major reason could be the optimization difficulty in larger and dense graphs, which could easily be stuck at one of many more stationary points. We leave it to future work to investigate these problems further.

\begin{table*}[!ht]\renewcommand{\arraystretch}{0.9}
\centering
\caption{Comparison of different algorithms on score differences from the ground truth, $\triangle F=F(\tilde{A},\mathbf{X})-F({A}^0,\mathbf{X})$. For each algorithm we show results as mean $\pm$ standard error over 100 trials.}
\label{Table:fullscore}
\vskip 0.1in
\begin{tabular}{c|cccc}
\hline
&$\lambda=10^2$ &$\lambda=(10,10^3)$ &NOTEARS&GOBNILP\\
\hline
ER3-Gaussian, $d=10$& $0.09\pm 0.20$&$0.06\pm0.20$&  $0.03\pm0.12$&$-0.03\pm0.00$\\
ER4-Gaussian, $d=10$& $0.14\pm0.03$ & $0.13\pm0.34$ & $0.08\pm0.21$&$-0.03\pm0.01$\\
ER6-Gaussian, $d=10$&$0.54\pm 0.22$&$0.36\pm0.75$& $0.22\pm0.40$&$-0.03\pm0.01$\\
SF4-Gumbel, $d=10$&{${-0.59\pm0.01}$}&$-0.59\pm0.14$ &${-0.71\pm0.08}$&$-1.73\pm0.07$\\
\hline
ER3-Gaussian, $d=30$& $0.33\pm0.19$&$0.07\pm0.04$ & $-0.06\pm0.02$&N/A\\
ER4-Gaussian, $d=30$& $0.31\pm0.05$&$0.40\pm0.19$ & $0.25\pm0.11$&N/A\\
ER6-Gaussian, $d=30$& $1.78\pm0.38$&$0.97\pm0.16$ & $1.02\pm0.18$&N/A\\
SF4-Gumbel, $d=30$&{$-3.30\pm0.04$}&$-3.31\pm0.02$ &$-3.55\pm0.02$&N/A\\
\hline
ER3-Gaussian, $d=50$& $0.05\pm0.05$&$-0.10\pm0.05$ & $-0.25\pm0.04$&N/A\\
ER4-Gaussian, $d=50$& $0.40\pm0.13$&$0.42\pm0.21$ & $0.19\pm0.09$&N/A\\
ER6-Gaussian, $d=50$& $2.31\pm0.41$&$1.77\pm0.38$ & $1.97\pm0.26$&N/A\\
SF4-Gumbel, $d=50$&{$-6.74\pm0.03$}&$-6.74\pm0.03$ &$-7.08\pm0.02$&N/A\\
\hline
ER3-Gaussian, $d=100$& $-0.82\pm0.16$&$-1.44\pm0.95$ & $-1.65\pm0.78$&N/A\\
ER4-Gaussian, $d=100$& $-0.28\pm0.26$&$-0.32\pm2.76$ & $-0.64\pm1.35$&N/A\\
ER6-Gaussian, $d=100$& $4.30\pm0.99$&$2.61\pm9.32$ & $2.49\pm 3.67$&N/A\\
SF4-Gumbel, $d=100$&{$-17.29\pm0.05$}&$-17.19\pm0.58$ &$-17.53\pm0.49$&N/A\\
\hline
\end{tabular}
\end{table*}

\begin{table*}[!ht]\renewcommand{\arraystretch}{0.9}
\centering
\caption{Hyperparameter and Ablation Study: results (mean $\pm$ standard error over 100 trials) for ER3-Gaussian Cases from DAG-NoCurl, where bold numbers highlight the best method for each case.}
\label{Table:hyperER3}
\vskip 0.1in
\scriptsize
\begin{tabular}{cccccccc}
\hline
$d$&Method&Time&$\Delta F$&SHD&\#Extra E&\#Missing E&\#Reverse E\\
\hline
10&$\lambda=1$&   0.08 $\pm$   0.00&   0.98 $\pm$   0.23&   4.12 $\pm$   0.30&   1.89 $\pm$   0.17&   1.50 $\pm$   0.14&   0.73 $\pm$   0.09\\
10&$\lambda=10$&  \bf{0.07 $\pm$   0.00}&   0.22 $\pm$   0.07&   1.50 $\pm$   0.22&   0.72 $\pm$   0.13&   0.31 $\pm$   0.08&   0.47 $\pm$   0.06\\
10&$\lambda=10^2$&   0.11 $\pm$   0.00&   0.09 $\pm$   0.02&   2.18 $\pm$   0.28&   1.24 $\pm$   0.18&   0.26 $\pm$   0.06&   0.68 $\pm$   0.08\\
10&$\lambda=10^3$&   0.38 $\pm$   0.01&   0.16 $\pm$   0.03&   3.14 $\pm$   0.37&   1.90 $\pm$   0.25&   0.39 $\pm$   0.08&   0.85 $\pm$   0.09\\
10&$\lambda=10^4$&   0.85 $\pm$   0.02&   0.28 $\pm$   0.04&   4.11 $\pm$   0.42&   2.43 $\pm$   0.28&   0.54 $\pm$   0.09&   1.14 $\pm$   0.10\\
10&$\lambda=(10,10^2)$&   0.23 $\pm$   0.01&   0.08 $\pm$   0.02&   1.24 $\pm$   0.20&   0.65 $\pm$   0.13&   0.19 $\pm$   0.05&  { 0.40 $\pm$   0.06}\\
10&$\lambda=(10,10^3)$&   0.47 $\pm$   0.01&   \textbf{0.06 $\pm$   0.02}&   \bf{1.08 $\pm$   0.18}&   {0.54 $\pm$   0.12}&   0.09 $\pm$   0.03&   0.45 $\pm$   0.06\\
10&$\lambda=(10,10^4)$&   0.85 $\pm$   0.02&   0.08 $\pm$   0.02&   1.43 $\pm$   0.23&   0.83 $\pm$   0.17&  \bf{ 0.08 $\pm$   0.03}&   0.52 $\pm$   0.07\\
10&$\lambda=(10^2,10^3)$&   0.44 $\pm$   0.01&   0.07 $\pm$   0.02&   2.09 $\pm$   0.27&   1.20 $\pm$   0.18&   0.25 $\pm$   0.06&   0.64 $\pm$   0.08\\
10&rand init&   0.19 $\pm$   0.01&   5.97 $\pm$   1.63&   9.73 $\pm$   0.54&   4.59 $\pm$   0.31&   2.78 $\pm$   0.26&   2.35 $\pm$   0.12\\
10&rand $p$&   0.32 $\pm$   0.03&  11.59 $\pm$   2.46&  14.88 $\pm$   0.57&   6.48 $\pm$   0.33&   5.06 $\pm$   0.30&   3.34 $\pm$   0.15\\
10&$\lambda=10^2$s&   0.09 $\pm$   0.00&   1.45 $\pm$   0.02&   3.11 $\pm$   0.31&   1.60 $\pm$   0.18&   0.95 $\pm$   0.11&   0.56 $\pm$   0.08\\
10&$\lambda=(10,10^3)$s&   0.43 $\pm$   0.01&   0.53 $\pm$   0.02&   1.27 $\pm$   0.20&   0.66 $\pm$   0.14&   0.20 $\pm$   0.05&   0.41 $\pm$   0.06\\
10&$\lambda=10^2$-&   0.11 $\pm$   0.00&  19.95 $\pm$   0.43&   3.62 $\pm$   0.27&   0.64 $\pm$   0.11&   2.93 $\pm$   0.19&   0.05 $\pm$   0.03\\
10&$\lambda=(10,10^3)$-&   0.32 $\pm$   0.01&  18.30 $\pm$   0.42&   2.09 $\pm$   0.17&   \textbf{0.30 $\pm$   0.08}&   1.79 $\pm$   0.13&   \textbf{0.00 $\pm$   0.00}\\
10&$\lambda=10^2$+&   0.29 $\pm$   0.01&   0.10 $\pm$   0.02&   2.15 $\pm$   0.27&   1.22 $\pm$   0.18&   0.25 $\pm$   0.06&   0.68 $\pm$   0.08\\
10&$\lambda=(10,10^3)$+&   0.71 $\pm$   0.01&   \textbf{0.06 $\pm$   0.02}&   \textbf{1.08 $\pm$   0.18}&   {0.54 $\pm$   0.12}&   0.09 $\pm$   0.03&   0.45 $\pm$   0.06\\
\hline
30&$\lambda=1$&  {0.54 $\pm$   0.03}&   2.55 $\pm$   0.37&  13.64 $\pm$   0.77&   8.34 $\pm$   0.55&   2.83 $\pm$   0.19&   2.47 $\pm$   0.15\\
30&$\lambda=10$&   0.59 $\pm$   0.03&   0.50 $\pm$   0.18&   6.46 $\pm$   0.50&   4.14 $\pm$   0.38&   0.62 $\pm$   0.09&   1.70 $\pm$   0.12\\
30&$\lambda=10^2$&   1.19 $\pm$   0.04&   0.33 $\pm$   0.19&   7.18 $\pm$   0.61&   5.05 $\pm$   0.49&   0.40 $\pm$   0.07&   1.73 $\pm$   0.12\\
30&$\lambda=10^3$&   1.29 $\pm$   0.03&   0.23 $\pm$   0.05&  10.13 $\pm$   0.72&   7.33 $\pm$   0.57&   0.41 $\pm$   0.07&   2.39 $\pm$   0.15\\
30&$\lambda=10^4$&   4.88 $\pm$   0.19&   0.23 $\pm$   0.04&  11.67 $\pm$   0.78&   8.36 $\pm$   0.60&   0.49 $\pm$   0.09&   2.82 $\pm$   0.16\\
30&$\lambda=(10,10^2)$&   0.64 $\pm$   0.02&   0.15 $\pm$   0.05&   5.41 $\pm$   0.49&   3.56 $\pm$   0.37&   0.42 $\pm$   0.08&   1.43 $\pm$   0.12\\
30&$\lambda=(10,10^3)$&   2.38 $\pm$   0.06&   0.07 $\pm$   0.04&   5.20 $\pm$   0.49&   3.63 $\pm$   0.39&   0.27 $\pm$   0.05&  { 1.30 $\pm$   0.10}\\
30&$\lambda=(10,10^4)$&   4.56 $\pm$   0.10&   \bf{0.00 $\pm$   0.02}&   \bf{4.92 $\pm$   0.45}&  { 3.38 $\pm$   0.38}&  \bf{ 0.14 $\pm$   0.04}&   1.40 $\pm$   0.09\\
30&$\lambda=(10^2,10^3)$&   2.50 $\pm$   0.06&   0.05 $\pm$   0.03&   6.61 $\pm$   0.63&   4.77 $\pm$   0.51&   0.24 $\pm$   0.05&   1.60 $\pm$   0.12\\
30&rand init&   2.39 $\pm$   0.14&   8.83 $\pm$   3.09&  30.96 $\pm$   1.37&  20.35 $\pm$   1.01&   6.05 $\pm$   0.43&   4.56 $\pm$   0.20\\
30&rand $p$&   3.40 $\pm$   0.23&  48.63 $\pm$   8.46&  59.57 $\pm$   1.51&  34.03 $\pm$   1.04&  16.71 $\pm$   0.59&   8.83 $\pm$   0.23\\
30&$\lambda=10^2$s&   {\bf0.29 $\pm$   0.01}&  17.27 $\pm$   0.20&  13.39 $\pm$   0.66&   8.50 $\pm$   0.53&   3.89 $\pm$   0.20&   1.00 $\pm$   0.10\\
30&$\lambda=(10,10^3)$s&   1.54 $\pm$   0.04&  10.98 $\pm$   0.19&   8.18 $\pm$   0.61&   5.26 $\pm$   0.47&   2.12 $\pm$   0.16&   0.80 $\pm$   0.08\\
30&$\lambda=10^2$-&   {0.34 $\pm$   0.01}& 104.65 $\pm$   1.09&  12.09 $\pm$   0.47&   1.94 $\pm$   0.22&  10.00 $\pm$   0.32&   0.15 $\pm$   0.04\\
30&$\lambda=(10,10^3)$-&   1.22 $\pm$   0.03&  95.39 $\pm$   1.04&   8.01 $\pm$   0.38&   \bf{1.34 $\pm$   0.17}&   6.60 $\pm$   0.28&   \bf{0.07 $\pm$   0.03}\\
30&$\lambda=10^2$+&   1.77 $\pm$   0.02&   0.32 $\pm$   0.19&   7.10 $\pm$   0.61&   5.00 $\pm$   0.49&   0.38 $\pm$   0.07&   1.72 $\pm$   0.12\\
30&$\lambda=(10,10^3)$+&   3.25 $\pm$   0.04&   0.07 $\pm$   0.04&   5.21 $\pm$   0.49&   3.64 $\pm$   0.39&   0.28 $\pm$   0.05&   1.29 $\pm$   0.10\\
\hline
50&$\lambda=1$&   2.73 $\pm$   0.14&   4.56 $\pm$   0.93&  25.16 $\pm$   1.23&  16.19 $\pm$   0.97&   4.33 $\pm$   0.25&   4.64 $\pm$   0.21\\
50&$\lambda=10$&  {2.00 $\pm$   0.10}&   0.38 $\pm$   0.09&  13.14 $\pm$   0.98&   9.16 $\pm$   0.79&   0.85 $\pm$   0.10&   3.13 $\pm$   0.19\\
50&$\lambda=10^2$&   2.32 $\pm$   0.09&   0.05 $\pm$   0.05&  13.51 $\pm$   1.00&   9.78 $\pm$   0.82&   0.65 $\pm$   0.10&   3.08 $\pm$   0.18\\
50&$\lambda=10^3$&   4.16 $\pm$   0.14&   0.03 $\pm$   0.05&  16.01 $\pm$   1.18&  11.60 $\pm$   0.96&   0.83 $\pm$   0.12&   3.58 $\pm$   0.18\\
50&$\lambda=10^4$&   8.40 $\pm$   0.26&   0.03 $\pm$   0.05&  18.95 $\pm$   1.08&  13.83 $\pm$   0.88&   0.67 $\pm$   0.10&   4.45 $\pm$   0.19\\
50&$\lambda=(10,10^2)$&   4.36 $\pm$   0.19&  -0.01 $\pm$   0.05&   9.95 $\pm$   0.79&   7.11 $\pm$   0.66&   0.60 $\pm$   0.07&   2.24 $\pm$   0.14\\
50&$\lambda=(10,10^3)$&   6.48 $\pm$   0.16&  -0.10 $\pm$   0.05&   {8.92 $\pm$   0.70}&   {6.35 $\pm$   0.57}&   0.41 $\pm$   0.08&  { 2.16 $\pm$   0.14}\\
50&$\lambda=(10,10^4)$&   8.73 $\pm$   0.20&  \bf{-0.16 $\pm$   0.06}&   9.21 $\pm$   0.69&   6.65 $\pm$   0.60&   \bf{0.21 $\pm$   0.05}&   2.35 $\pm$   0.14\\
50&$\lambda=(10^2,10^3)$&   5.20 $\pm$   0.16&  -0.01 $\pm$   0.06&  11.98 $\pm$   0.96&   8.69 $\pm$   0.78&   0.49 $\pm$   0.09&   2.80 $\pm$   0.17\\
50&rand init&   8.48 $\pm$   0.56&  12.03 $\pm$   0.65&  72.31 $\pm$   2.55&  52.15 $\pm$   2.08&  14.26 $\pm$   0.71&   5.91 $\pm$   0.27\\
50&rand $p$&  11.77 $\pm$   0.60&  78.54 $\pm$   9.59& 112.86 $\pm$   2.38&  68.74 $\pm$   1.72&  29.64 $\pm$   0.85&  14.48 $\pm$   0.30\\
50&$\lambda=10^2$s&   \bf{1.37 $\pm$   0.06}&  48.15 $\pm$   1.02&  25.53 $\pm$   1.08&  16.85 $\pm$   0.81&   7.02 $\pm$   0.32&   1.66 $\pm$   0.12\\
50&$\lambda=(10,10^3)$s&   4.30 $\pm$   0.10&  24.66 $\pm$   0.40&  15.77 $\pm$   0.71&  10.77 $\pm$   0.57&   3.80 $\pm$   0.19&   1.20 $\pm$   0.10\\
50&$\lambda=10^2$-&   2.04 $\pm$   0.08& 272.99 $\pm$   4.79&  20.45 $\pm$   0.69&   2.97 $\pm$   0.28&  17.31 $\pm$   0.48&   0.17 $\pm$   0.04\\
50&$\lambda=(10,10^3)$-&   4.65 $\pm$   0.11& 252.60 $\pm$   4.65&  13.53 $\pm$   0.47&   \bf{2.19 $\pm$   0.19}&  11.25 $\pm$   0.37&   \bf{0.09 $\pm$   0.03}\\
50&$\lambda=10^2$+&   3.13 $\pm$   0.06&   0.05 $\pm$   0.05&  13.63 $\pm$   1.01&   9.87 $\pm$   0.82&   0.66 $\pm$   0.10&   3.10 $\pm$   0.18\\
50&$\lambda=(10,10^3)$+&   9.36 $\pm$   0.10&  -0.11 $\pm$   0.05&   \bf{8.88 $\pm$   0.70}&   6.33 $\pm$   0.57&   0.39 $\pm$   0.08&   2.16 $\pm$   0.15\\
\hline
100&$\lambda=1$&   {8.05 $\pm$   0.40}&   6.36 $\pm$   0.62&  54.27 $\pm$   1.87&  36.22 $\pm$   1.49&   8.69 $\pm$   0.31&   9.36 $\pm$   0.28\\
100&$\lambda=10$&  16.02 $\pm$   0.75&   0.50 $\pm$   0.27&  29.14 $\pm$   1.44&  20.79 $\pm$   1.17&   1.91 $\pm$   0.16&   6.44 $\pm$   0.26\\
100&$\lambda=10^2$&  18.20 $\pm$   0.81&  -0.82 $\pm$   0.16&  31.99 $\pm$   1.66&  24.09 $\pm$   1.32&   1.30 $\pm$   0.16&   6.60 $\pm$   0.30\\
100&$\lambda=10^3$&  32.59 $\pm$   0.90&  -1.05 $\pm$   0.12&  37.99 $\pm$   1.74&  28.91 $\pm$   1.46&   1.25 $\pm$   0.15&   7.83 $\pm$   0.29\\
100&$\lambda=10^4$&  56.89 $\pm$   1.48&  -1.04 $\pm$   0.16&  43.16 $\pm$   1.72&  32.92 $\pm$   1.44&   0.98 $\pm$   0.12&   9.26 $\pm$   0.29\\
100&$\lambda=(10,10^2)$&  12.48 $\pm$   0.54&  -0.92 $\pm$   0.12&  23.48 $\pm$   1.37&  17.36 $\pm$   1.14&   0.99 $\pm$   0.10&   5.13 $\pm$   0.23\\
100&$\lambda=(10,10^3)$&  26.02 $\pm$   0.76&  -1.44 $\pm$   0.09& \bf{ 19.16 $\pm$   1.10}&  {14.30 $\pm$   0.89}&   0.62 $\pm$   0.09&   {4.24 $\pm$   0.23}\\
100&$\lambda=(10,10^4)$&  62.12 $\pm$   1.69&  \bf{-1.52 $\pm$   0.10}&  19.66 $\pm$   1.24&  14.70 $\pm$   1.03&  \bf{ 0.38 $\pm$   0.09}&   4.58 $\pm$   0.23\\
100&$\lambda=(10^2,10^3)$&  26.69 $\pm$   0.76&  -1.20 $\pm$   0.15&  27.81 $\pm$   1.49&  21.08 $\pm$   1.21&   0.98 $\pm$   0.14&   5.75 $\pm$   0.27\\
100&rand init&  35.48 $\pm$   1.67&  25.53 $\pm$   1.82& 171.88 $\pm$   4.79& 129.30 $\pm$   3.80&  32.63 $\pm$   1.45&   9.94 $\pm$   0.41\\
100&rand $p$&  49.62 $\pm$   1.77& 151.17 $\pm$  19.44& 247.30 $\pm$   4.09& 159.63 $\pm$   3.25&  59.22 $\pm$   1.16&  28.45 $\pm$   0.41\\
100&$\lambda=10^2$s&   \bf{6.55 $\pm$   0.30}& 100.83 $\pm$   0.84&  57.03 $\pm$   1.49&  38.63 $\pm$   1.20&  14.85 $\pm$   0.38&   3.55 $\pm$   0.20\\
100&$\lambda=(10,10^3)$s&  21.11 $\pm$   0.56&  66.03 $\pm$   0.66&  35.75 $\pm$   1.17&  25.47 $\pm$   0.95&   7.96 $\pm$   0.29&   2.32 $\pm$   0.15\\
100&$\lambda=10^2$-&   {7.65 $\pm$   0.36}& 480.18 $\pm$   4.61&  45.27 $\pm$   1.00&   7.37 $\pm$   0.41&  37.48 $\pm$   0.68&   0.42 $\pm$   0.06\\
100&$\lambda=(10,10^3)$-&  19.40 $\pm$   0.51& 440.55 $\pm$   4.42&  29.21 $\pm$   0.74&   \bf{4.52 $\pm$   0.29}&  24.53 $\pm$   0.57&   \bf{0.16 $\pm$   0.04}\\
100&$\lambda=10^2$+&   44.8 $\pm$   0.44&  -0.84 $\pm$   0.16&  31.88 $\pm$   1.66&  24.00 $\pm$   1.32&   1.29 $\pm$   0.16&   6.59 $\pm$   0.30\\
100&$\lambda=(10,10^3)$+&  57.9 $\pm$   0.50&  -1.43 $\pm$   0.10&  19.30 $\pm$   1.14&  14.42 $\pm$   0.92&   0.60 $\pm$   0.09&   4.28 $\pm$   0.23\\
\hline
\end{tabular}
\end{table*}

\section{Detailed Results for Structure Recovery}\label{sec:SHD}

In this section we provide the detailed numerical results of linear synthetic datasets for different algorithms, as a continuation of the discussion in Section \ref{sec:structure}  of the main text and as the supplementary results of the structure discovery in terms of SHD and the run time plotted in Figure \ref{fig:nocurl_linear} of the main text. The full results for ER3-Gaussian, ER4-Gaussian, ER6-Gaussian and SF4-Gumbel cases are provided in Tables \ref{Table:score1}, \ref{Table:score2}, \ref{Table:score3} and \ref{Table:score4}, respectively. Besides SHD, we further list $\Delta F$, the number of extra edges, missing edges and reverse edges as additional algorithm evaluation metric. From these tables we can see that the most accurate structure discovery results in terms of SHD are either from NOTEARS or NoCurl, while the other three algorithms (FGS, CAM and MMPC) rapidly deteriorates as the number of edges increase. Among the total $16$ cases with different combinations of $d\in\{10,30,50,100\}$ and graph/noise-types, NoCurl-2 outperforms NOTEARS (as well as all other algorithms) with a lower SHD in most (12 out of 16) cases. We further observe that the low SHD from NoCurl comes from the fact that this algorithm tends to miss much fewer numbers of edges comparing with other algorithms especially in large and dense graphs, possibly because Step 2 in NoCurl has successfully recovered some lost edges, as we have observed and discussed in the Ablation Study section \ref{sec:ablation} above. When comparing the computational time, NoCurl is faster than NOTEARS by one or two orders of magnitude.

\begin{table*}[!ht]\renewcommand{\arraystretch}{0.9}
\centering
\caption{Hyperparameter and Ablation Study: results (mean $\pm$ standard error over 100 trials) for ER6-Gaussian Cases from DAG-NoCurl, where bold numbers highlight the best method for each case.}
\label{Table:hyperER6}
\vskip 0.1in
\scriptsize
\begin{tabular}{cccccccc}
\hline
$d$&Method&Time&$\Delta F$&SHD&\#Extra E&\#Missing E&\#Reverse E\\
\hline
10&$\lambda=1$&  {0.33 $\pm$   0.02}&   2.49 $\pm$   0.27&   7.08 $\pm$   0.43&   2.57 $\pm$   0.23&   3.27 $\pm$   0.22&   1.24 $\pm$   0.09\\
10&$\lambda=10$&   0.37 $\pm$   0.02&   0.83 $\pm$   0.14&   3.46 $\pm$   0.33&   1.18 $\pm$   0.14&   1.36 $\pm$   0.16&   0.92 $\pm$   0.09\\
10&$\lambda=10^2$&   0.34 $\pm$   0.02&   0.54 $\pm$   0.22&   3.54 $\pm$   0.37&   1.37 $\pm$   0.18&   1.30 $\pm$   0.17&   0.87 $\pm$   0.08\\
10&$\lambda=10^3$&   0.56 $\pm$   0.02&   0.48 $\pm$   0.05&   5.74 $\pm$   0.46&   2.58 $\pm$   0.26&   1.64 $\pm$   0.18&   1.52 $\pm$   0.11\\
10&$\lambda=10^4$&   1.16 $\pm$   0.03&   0.75 $\pm$   0.08&   6.98 $\pm$   0.51&   3.15 $\pm$   0.28&   2.07 $\pm$   0.20&   1.76 $\pm$   0.12\\
10&$\lambda=(10,10^2)$&   0.62 $\pm$   0.03&   0.76 $\pm$   0.21&   3.10 $\pm$   0.32&   0.99 $\pm$   0.13&   1.24 $\pm$   0.16&   0.87 $\pm$   0.09\\
10&$\lambda=(10,10^3)$&   0.75 $\pm$   0.03&   \bf{ 0.36 $\pm$   0.07}&   3.07 $\pm$   0.30&   { 0.97 $\pm$   0.13}&   1.02 $\pm$   0.14&   1.08 $\pm$   0.09\\
10&$\lambda=(10,10^4)$&   1.25 $\pm$   0.04&   0.49 $\pm$   0.09&  \bf{ 2.97 $\pm$   0.32}&   0.99 $\pm$   0.13&   \bf{ 0.88 $\pm$   0.14}&   1.10 $\pm$   0.10\\
10&$\lambda=(10^2,10^3)$&   0.72 $\pm$   0.03&   0.37 $\pm$   0.17&   3.29 $\pm$   0.35&   1.31 $\pm$   0.18&   1.01 $\pm$   0.15&   0.97 $\pm$   0.09\\
10&rand init&   0.37 $\pm$   0.03&  78.00 $\pm$  40.34&  17.09 $\pm$   0.62&   5.63 $\pm$   0.26&   8.20 $\pm$   0.53&   3.25 $\pm$   0.15\\
10&rand $p$&   0.66 $\pm$   0.06& 128.79 $\pm$  47.97&  24.82 $\pm$   0.47&   6.10 $\pm$   0.22&  13.84 $\pm$   0.41&   4.88 $\pm$   0.20\\
10&$\lambda=10^2$s&   0.37 $\pm$   0.03&   8.23 $\pm$   0.15&   4.79 $\pm$   0.37&   1.78 $\pm$   0.18&   2.32 $\pm$   0.23&   0.69 $\pm$   0.08\\
10&$\lambda=(10,10^3)$s&   0.92 $\pm$   0.03&   5.55 $\pm$   0.32&   3.27 $\pm$   0.35&   1.10 $\pm$   0.16&   1.19 $\pm$   0.17&   0.98 $\pm$   0.09\\
10&$\lambda=10^2$-&   \bf{0.21 $\pm$   0.01}& 339.60 $\pm$  10.48&   7.28 $\pm$   0.42&   0.75 $\pm$   0.11&   6.52 $\pm$   0.35&   \bf{0.01 $\pm$   0.01}\\
10&$\lambda=(10,10^3)$-&   0.69 $\pm$   0.03& 332.83 $\pm$  10.62&   5.19 $\pm$   0.39&   \bf{0.53 $\pm$   0.09}&   4.64 $\pm$   0.33&   0.02 $\pm$   0.01\\
10&$\lambda=10^2$+&   0.53 $\pm$   0.02&   0.54 $\pm$   0.22&   3.54 $\pm$   0.37&   1.37 $\pm$   0.18&   1.30 $\pm$   0.17&  {  0.87 $\pm$   0.08}\\
10&$\lambda=(10,10^3)$+&   1.45 $\pm$   0.04&   \bf{0.36 $\pm$   0.07}&   3.07 $\pm$   0.30&   {0.97 $\pm$   0.13}&   1.02 $\pm$   0.14&   1.08 $\pm$   0.09\\
\hline
30&$\lambda=1$&   2.63 $\pm$   0.17&  16.64 $\pm$   3.57&  46.90 $\pm$   1.76&  32.35 $\pm$   1.34&   8.46 $\pm$   0.46&   6.09 $\pm$   0.22\\
30&$\lambda=10$&   3.26 $\pm$   0.19&   4.22 $\pm$   0.68&  27.34 $\pm$   1.91&  19.59 $\pm$   1.49&   3.53 $\pm$   0.36&   4.22 $\pm$   0.20\\
30&$\lambda=10^2$&   3.34 $\pm$   0.21&   1.78 $\pm$   0.38&  21.44 $\pm$   1.56&  15.98 $\pm$   1.16&   2.41 $\pm$   0.33&   3.05 $\pm$   0.19\\
30&$\lambda=10^3$&   {2.71 $\pm$   0.13}&   1.42 $\pm$   0.32&  24.23 $\pm$   1.81&  18.38 $\pm$   1.38&   2.61 $\pm$   0.34&   3.24 $\pm$   0.18\\
30&$\lambda=10^4$&   6.56 $\pm$   0.20&   1.25 $\pm$   0.31&  30.03 $\pm$   1.73&  23.19 $\pm$   1.35&   2.78 $\pm$   0.28&   4.06 $\pm$   0.20\\
30&$\lambda=(10,10^2)$&   5.03 $\pm$   0.29&   2.09 $\pm$   0.57&  19.99 $\pm$   1.50&  14.77 $\pm$   1.18&   2.21 $\pm$   0.27&   3.01 $\pm$   0.19\\
30&$\lambda=(10,10^3)$&   7.68 $\pm$   0.39&   \bf{ 0.97 $\pm$   0.16}&  \bf{17.37 $\pm$   1.18}&  12.93 $\pm$   0.92&   \bf{1.71 $\pm$   0.19}&   {2.73 $\pm$   0.16}\\
30&$\lambda=(10,10^4)$&   8.73 $\pm$   0.29&   0.98 $\pm$   0.20&  17.39 $\pm$   1.44&  {12.79 $\pm$   1.11}&   \bf{1.71 $\pm$   0.26}&   2.89 $\pm$   0.18\\
30&$\lambda=(10^2,10^3)$&   3.96 $\pm$   0.21&   1.23 $\pm$   0.34&  20.19 $\pm$   1.71&  15.29 $\pm$   1.32&   2.08 $\pm$   0.30&   2.82 $\pm$   0.19\\
30&rand init&   6.04 $\pm$   0.26&  41.38 $\pm$  21.75&  84.88 $\pm$   2.65&  59.53 $\pm$   1.94&  18.03 $\pm$   0.85&   7.32 $\pm$   0.21\\
30&rand $p$&   7.74 $\pm$   0.38& 871.38 $\pm$ 198.66& 130.37 $\pm$   1.66&  70.36 $\pm$   1.26&  49.65 $\pm$   1.00&  10.36 $\pm$   0.28\\
30&$\lambda=10^2$s&   \bf{1.58 $\pm$   0.10}& 824.91 $\pm$  15.39&  37.36 $\pm$   1.34&  24.83 $\pm$   0.94&  11.04 $\pm$   0.47&   1.49 $\pm$   0.14\\
30&$\lambda=(10,10^3)$s&   4.69 $\pm$   0.23& 667.06 $\pm$  17.54&  29.88 $\pm$   1.55&  20.25 $\pm$   1.12&   8.01 $\pm$   0.46&   1.62 $\pm$   0.13\\
30&$\lambda=10^2$-&   
{1.69 $\pm$   0.11}& 4908.93 $\pm$  96.30&  32.82 $\pm$   1.07&   \bf{5.11 $\pm$   0.40}&  27.51 $\pm$   0.77&   0.20 $\pm$   0.05\\
30&$\lambda=(10,10^3)$-&   4.67 $\pm$   0.23& 4610.52 $\pm$  95.17&  26.08 $\pm$   1.07&   5.42 $\pm$   0.47&  20.51 $\pm$   0.70&   \bf{0.15 $\pm$   0.04}\\
30&$\lambda=10^2$+&   5.32 $\pm$   0.35&   1.78 $\pm$   0.38&  21.44 $\pm$   1.56&  15.98 $\pm$   1.16&   2.41 $\pm$   0.33&   3.05 $\pm$   0.19\\
30&$\lambda=(10,10^3)$+&   10.38 $\pm$   0.23&   1.02 $\pm$   0.22&  17.81 $\pm$   1.29&  13.37 $\pm$   1.04&   1.74 $\pm$   0.18&   2.70 $\pm$   0.16\\
\hline
50&$\lambda=1$&  14.74 $\pm$   0.85&  54.63 $\pm$  16.95&  95.58 $\pm$   3.36&  70.11 $\pm$   2.52&  15.54 $\pm$   0.91&   9.93 $\pm$   0.28\\
50&$\lambda=10$&  15.22 $\pm$   0.90&   6.71 $\pm$   0.81&  55.14 $\pm$   2.88&  42.10 $\pm$   2.40&   5.57 $\pm$   0.43&   7.47 $\pm$   0.23\\
50&$\lambda=10^2$&  12.05 $\pm$   0.77&   2.31 $\pm$   0.41&  40.32 $\pm$   2.40&  32.10 $\pm$   2.00&   2.83 $\pm$   0.24&   5.39 $\pm$   0.26\\
50&$\lambda=10^3$&  { 10.13 $\pm$   0.56}&   1.98 $\pm$   0.73&  40.61 $\pm$   2.96&  32.65 $\pm$   2.34&   2.93 $\pm$   0.49&   5.03 $\pm$   0.27\\
50&$\lambda=10^4$&  19.66 $\pm$   0.62&   \bf{ 1.01 $\pm$   0.38}&  43.44 $\pm$   2.79&  34.93 $\pm$   2.26&   2.95 $\pm$   0.44&   5.56 $\pm$   0.22\\
50&$\lambda=(10,10^2)$&  29.23 $\pm$   1.75&   3.10 $\pm$   0.50&  35.92 $\pm$   2.16&  28.07 $\pm$   1.88&   2.75 $\pm$   0.23&   5.10 $\pm$   0.19\\
50&$\lambda=(10,10^3)$&  31.74 $\pm$   1.71&   1.77 $\pm$   0.38&  33.67 $\pm$   2.53&  26.69 $\pm$   2.08&   2.45 $\pm$   0.35&   4.53 $\pm$   0.22\\
50&$\lambda=(10,10^4)$&  24.14 $\pm$   0.85&   1.15 $\pm$   0.27&  \bf{ 32.27 $\pm$   2.44}&  { 25.43 $\pm$   2.02}&   \bf{ 2.06 $\pm$   0.34}&   4.78 $\pm$   0.22\\
50&$\lambda=(10^2,10^3)$&  15.50 $\pm$   0.84&   2.16 $\pm$   0.63&  34.87 $\pm$   2.56&  28.06 $\pm$   2.05&   2.50 $\pm$   0.38&   {4.31 $\pm$   0.26}\\
50&rand init&  17.37 $\pm$   0.34&  47.87 $\pm$  27.87& 156.03 $\pm$   5.19& 118.60 $\pm$   4.16&  26.79 $\pm$   1.15&  10.64 $\pm$   0.26\\
50&rand $p$&  24.40 $\pm$   1.03& 1497.50 $\pm$ 407.17& 255.90 $\pm$   3.06& 155.04 $\pm$   2.35&  85.71 $\pm$   1.51&  15.14 $\pm$   0.33\\
50&$\lambda=10^2$s&   8.45 $\pm$   0.55& 2957.04 $\pm$  69.41&  69.24 $\pm$   2.00&  48.31 $\pm$   1.57&  19.01 $\pm$   0.54&   1.92 $\pm$   0.15\\
50&$\lambda=(10,10^3)$s&  12.51 $\pm$   0.68& 2114.01 $\pm$  51.61&  56.68 $\pm$   1.95&  40.93 $\pm$   1.55&  13.62 $\pm$   0.52&   2.13 $\pm$   0.14\\
50&$\lambda=10^2$-&   \bf{5.39 $\pm$   0.34}& 16135.01 $\pm$ 389.36&  56.18 $\pm$   1.46&   \bf{8.10 $\pm$   0.58}&  47.85 $\pm$   1.05&   0.23 $\pm$   0.05\\
50&$\lambda=(10,10^3)$-&  13.94 $\pm$   0.76& 15500.46 $\pm$ 378.96&  44.66 $\pm$   1.37&   8.97 $\pm$   0.57&  35.54 $\pm$   0.95&   \bf{0.15 $\pm$   0.04}\\
50&$\lambda=10^2$+&  18.03 $\pm$   0.78&   2.31 $\pm$   0.41&  40.31 $\pm$   2.39&  32.10 $\pm$   2.00&   2.82 $\pm$   0.24&   5.39 $\pm$   0.26\\
50&$\lambda=(10,10^3)$+&  44.13 $\pm$   0.81&   1.78 $\pm$   0.46&  33.41 $\pm$   2.63&  26.40 $\pm$   2.15&   2.56 $\pm$   0.41&   4.45 $\pm$   0.21\\
\hline
100&$\lambda=1$&  52.40 $\pm$   1.68&  67.18 $\pm$   6.59& 215.70 $\pm$   6.90& 165.76 $\pm$   5.56&  30.96 $\pm$   1.38&  18.98 $\pm$   0.40\\
100&$\lambda=10$&  62.17 $\pm$   2.19&  20.28 $\pm$   4.99& 122.05 $\pm$   4.26&  96.67 $\pm$   3.62&  10.23 $\pm$   0.56&  15.15 $\pm$   0.35\\
100&$\lambda=10^2$&  {41.29 $\pm$   1.56}&   4.30 $\pm$   0.99&  89.93 $\pm$   3.49&  74.34 $\pm$   3.04&   4.64 $\pm$   0.34&  10.95 $\pm$   0.33\\
100&$\lambda=10^3$&  60.88 $\pm$   2.30&   1.98 $\pm$   0.86&  92.12 $\pm$   4.32&  77.68 $\pm$   3.69&   4.21 $\pm$   0.48&  10.23 $\pm$   0.32\\
100&$\lambda=10^4$&  94.68 $\pm$   1.99&  \bf{-0.03 $\pm$   0.22}&  92.36 $\pm$   3.36&  78.60 $\pm$   2.97&   3.28 $\pm$   0.30&  10.48 $\pm$   0.27\\
100&$\lambda=(10,10^2)$& 152.84 $\pm$   6.00&   4.68 $\pm$   0.77&  84.02 $\pm$   3.58&  68.57 $\pm$   3.10&   4.91 $\pm$   0.39&  10.54 $\pm$   0.33\\
100&$\lambda=(10,10^3)$&  84.24 $\pm$   3.26&   2.61 $\pm$   0.93&  72.30 $\pm$   3.80&  60.12 $\pm$   3.49&   3.58 $\pm$   0.30&   8.60 $\pm$   0.27\\
100&$\lambda=(10,10^4)$& 135.50 $\pm$   3.39&   0.88 $\pm$   0.32& \bf{66.13 $\pm$   2.85}&  {54.66 $\pm$   2.53}&  \bf{2.92 $\pm$   0.27}&   {8.55 $\pm$   0.26}\\
100&$\lambda=(10^2,10^3)$& 146.64 $\pm$   5.87&   2.64 $\pm$   1.08&  85.73 $\pm$   5.71&  72.07 $\pm$   4.72&   4.66 $\pm$   0.83&   9.00 $\pm$   0.33\\
100&rand init&  51.32 $\pm$   0.31&  49.17 $\pm$   6.88& 445.43 $\pm$  16.00& 367.63 $\pm$  14.03&  61.00 $\pm$   2.32&  16.80 $\pm$   0.37\\
100&rand $p$&  67.64 $\pm$   2.14& 3895.88 $\pm$ 1085.90& 608.89 $\pm$   6.27& 409.29 $\pm$   5.61& 173.17 $\pm$   2.24&  26.43 $\pm$   0.47\\
100&$\lambda=10^2$s&  43.98 $\pm$   2.11& 6319.57 $\pm$ 103.73& 138.70 $\pm$   2.61&  97.83 $\pm$   2.14&  37.22 $\pm$   0.69&   3.65 $\pm$   0.16\\
100&$\lambda=(10,10^3)$s& 106.88 $\pm$   4.64& 5285.84 $\pm$ 124.45& 108.05 $\pm$   2.74&  78.33 $\pm$   2.27&  26.03 $\pm$   0.60&   3.69 $\pm$   0.19\\
100&$\lambda=10^2$-&  \bf{32.56 $\pm$   1.53}& 31457.27 $\pm$ 604.87& 110.61 $\pm$   1.75&  14.13 $\pm$   0.66&  96.12 $\pm$   1.36&   0.36 $\pm$   0.07\\
100&$\lambda=(10,10^3)$-& 121.89 $\pm$   5.33& 30154.20 $\pm$ 608.15&  83.21 $\pm$   1.61&  \bf{13.67 $\pm$   0.75}&  69.26 $\pm$   1.12&   \bf{0.28 $\pm$   0.06}\\
100&$\lambda=10^2$+&  84.24 $\pm$   3.38&   4.04 $\pm$   0.93&  89.64 $\pm$   3.47&  74.27 $\pm$   3.04&   4.41 $\pm$   0.34&  10.96 $\pm$   0.31\\
100&$\lambda=(10,10^3)$+&  107.87 $\pm$   3.07&   2.60 $\pm$   0.93&  72.35 $\pm$   3.81&  60.17 $\pm$   3.49&   3.58 $\pm$   0.30&   8.60 $\pm$   0.27\\
\hline
\end{tabular}
\end{table*}

\begin{table*}[!ht]\renewcommand{\arraystretch}{0.9}
\centering
\caption{Comparison of Different Algorithms on Linear Synthetic Datasets: results (mean $\pm$ standard error over 100 trials) for ER3-Gaussian Cases, where bold numbers highlight the best method for each case.}
\label{Table:score1}
\vskip 0.1in
\small
\begin{tabular}{cccccccc}
\hline
$d$&Method&Time&$\Delta F$&SHD&\#Extra E&\#Missing E&\#Reverse E\\
\hline
10&NOTEARS&   1.71 $\pm$   0.07&   \bf{0.03 $\pm$   0.01}&   1.11 $\pm$   0.21&   0.55 $\pm$   0.14&   0.15 $\pm$   0.05&   {0.41 $\pm$   0.06}\\
10&FGS&   0.65 $\pm$   0.07&   --&   6.34 $\pm$   0.55&   2.85 $\pm$   0.37&   0.98 $\pm$   0.13&   2.51 $\pm$   0.18\\
10&CAM&   8.46 $\pm$   0.16&   --&  12.34 $\pm$   0.61&   5.05 $\pm$   0.34&   1.77 $\pm$   0.17&   5.52 $\pm$   0.23\\
10&MMPC&   0.89 $\pm$   0.03&   --&  15.36 $\pm$   0.36&   0.68 $\pm$   0.09&   3.78 $\pm$   0.30&  10.90 $\pm$   0.15\\
10&Eq+BU&   0.57 $\pm$   0.01& -- &   2.92 $\pm$   0.21&   2.91 $\pm$   0.21&   \bf{0.01 $\pm$   0.01}&   \bf{0.00 $\pm$   0.00}\\
10&Eq+TD&   0.58 $\pm$   0.02& --&   3.21 $\pm$   0.23&   3.20 $\pm$   0.23&   \bf{0.01 $\pm$   0.01}&   \bf{0.00 $\pm$   0.00}\\
10&NoCurl-1{\color{red}s}&   \bf{0.09 $\pm$   0.00}&   1.45 $\pm$   0.02&   3.11 $\pm$   0.31&   1.60 $\pm$   0.18&   0.95 $\pm$   0.11&   0.56 $\pm$   0.08\\
10&NoCurl-2{\color{red}s}&   0.43 $\pm$   0.01&   0.53 $\pm$   0.02&   1.27 $\pm$   0.20&   0.66 $\pm$   0.14&   0.20 $\pm$   0.05&   {0.41 $\pm$   0.06}\\
10&NoCurl-1&   {0.11 $\pm$   0.00}&   0.09 $\pm$   0.02&   2.18 $\pm$   0.28&   1.24 $\pm$   0.18&   0.26 $\pm$   0.06&   0.68 $\pm$   0.08\\
10&NoCurl-2&   0.47 $\pm$   0.01&   0.06 $\pm$   0.02&   \bf{1.08 $\pm$   0.18}&   \bf{0.54 $\pm$   0.12}&   {0.09 $\pm$   0.03}&   0.45 $\pm$   0.06\\
\hline
30&NOTEARS&  37.25 $\pm$   1.67&  \bf{-0.06 $\pm$   0.02}&   \bf{4.42 $\pm$   0.48}&   2.85 $\pm$   0.36&   0.47 $\pm$   0.11&   {1.10 $\pm$   0.10}\\
30&FGS&   {0.96 $\pm$   0.04}&   --&  15.16 $\pm$   1.33&   8.53 $\pm$   1.05&   1.98 $\pm$   0.23&   4.65 $\pm$   0.24\\
30&CAM&  45.87 $\pm$   0.94&   --&  36.27 $\pm$   1.17&  18.23 $\pm$   0.70&   4.34 $\pm$   0.31&  13.70 $\pm$   0.37\\
30&MMPC&   1.74 $\pm$   0.05&   --&  46.67 $\pm$   0.68&   \bf{2.62 $\pm$   0.18}&  11.72 $\pm$   0.60&  32.33 $\pm$   0.34\\
30&Eq+BU&   2.12 $\pm$   0.01& --&  14.14 $\pm$   0.75&  14.12 $\pm$   0.75&   \bf{0.02 $\pm$   0.01}&   \bf{0.00 $\pm$   0.00}\\
30&Eq+TD&   2.07 $\pm$   0.01& --&  15.45 $\pm$   0.82&  15.43 $\pm$   0.81&   \bf{0.02 $\pm$   0.01}&   \bf{0.00 $\pm$   0.00}\\
30&NoCurl-1{\color{red}s}&   \bf{0.29 $\pm$   0.01}&  17.27 $\pm$   0.20&  13.39 $\pm$   0.66&   8.50 $\pm$   0.53&   3.89 $\pm$   0.20&   1.00 $\pm$   0.10\\
30&NoCurl-2{\color{red}s}&   1.54 $\pm$   0.04&  10.98 $\pm$   0.19&   8.18 $\pm$   0.61&   5.26 $\pm$   0.47&   2.12 $\pm$   0.16&   0.80 $\pm$   0.08\\
30&NoCurl-1&   1.19 $\pm$   0.04&   0.33 $\pm$   0.19&   7.18 $\pm$   0.61&   5.05 $\pm$   0.49&   0.40 $\pm$   0.07&   1.73 $\pm$   0.12\\
30&NoCurl-2&   2.38 $\pm$   0.06&   0.07 $\pm$   0.04&   5.20 $\pm$   0.49&   3.63 $\pm$   0.39&   {0.27 $\pm$   0.05}&   1.30 $\pm$   0.10\\
\hline
50&NOTEARS& 253.96 $\pm$   9.49&  \bf{-0.25 $\pm$   0.04}&   \bf{8.39 $\pm$   0.70}&   5.56 $\pm$   0.53&   1.24 $\pm$   0.18&   {1.59 $\pm$   0.13}\\
50&FGS&   {1.42 $\pm$   0.06}&   --&  26.90 $\pm$   2.14&  16.21 $\pm$   1.85&   3.48 $\pm$   0.30&   7.21 $\pm$   0.26\\
50&CAM&  75.11 $\pm$   0.73&   --&  59.03 $\pm$   1.58&  29.31 $\pm$   0.97&   7.71 $\pm$   0.39&  22.01 $\pm$   0.47\\
50&MMPC&   3.88 $\pm$   0.11&   --&  78.82 $\pm$   0.86&   \bf{4.32 $\pm$   0.23}&  19.39 $\pm$   0.68&  55.11 $\pm$   0.50\\
50&Eq+BU&   4.59 $\pm$   0.05& --&  27.06 $\pm$   1.12&  26.99 $\pm$   1.12&   0.07 $\pm$   0.04&   \bf{0.00 $\pm$   0.00}\\
50&Eq+TD&   4.29 $\pm$   0.05& --&  29.39 $\pm$   1.24&  29.33 $\pm$   1.23&   \bf{0.06 $\pm$   0.04}&   \bf{0.00 $\pm$   0.00}\\
50&NoCurl-1{\color{red}s}&   \bf{1.37 $\pm$   0.06}&  48.15 $\pm$   1.02&  25.53 $\pm$   1.08&  16.85 $\pm$   0.81&   7.02 $\pm$   0.32&   1.66 $\pm$   0.12\\
50&NoCurl-2{\color{red}s}&   4.30 $\pm$   0.10&  24.66 $\pm$   0.40&  15.77 $\pm$   0.71&  10.77 $\pm$   0.57&   3.80 $\pm$   0.19&   1.20 $\pm$   0.10\\
50&NoCurl-1&   2.32 $\pm$   0.09&   0.05 $\pm$   0.05&  13.51 $\pm$   1.00&   9.78 $\pm$   0.82&   0.65 $\pm$   0.10&   3.08 $\pm$   0.18\\
50&NoCurl-2&   6.48 $\pm$   0.16&  -0.10 $\pm$   0.05&   8.92 $\pm$   0.70&   6.35 $\pm$   0.57&   {0.41 $\pm$   0.08}&   2.16 $\pm$   0.14\\
\hline
100&NOTEARS& 659.35 $\pm$  10.91&  \bf{-1.65 $\pm$   0.08}&  22.26 $\pm$   1.58&  16.28 $\pm$   1.22&   3.77 $\pm$   0.35&   {2.21 $\pm$   0.14}\\
100&FGS&   \bf{2.36 $\pm$   0.09}&   --&  34.12 $\pm$   2.04&  16.16 $\pm$   1.69&   5.54 $\pm$   0.38&  12.42 $\pm$   0.37\\
100&CAM& 197.76 $\pm$   1.89&   --& 104.99 $\pm$   2.00&  51.07 $\pm$   1.24&  12.21 $\pm$   0.55&  41.71 $\pm$   0.68\\
100&MMPC&   6.38 $\pm$   0.16&   --& 159.40 $\pm$   1.19&  \bf{10.00 $\pm$   0.38}&  32.39 $\pm$   1.19& 117.01 $\pm$   0.84\\
100&Eq+BU&  15.31 $\pm$   0.14& --&  52.94 $\pm$   2.28&  52.84 $\pm$   2.27&   \bf{0.10 $\pm$   0.05}&   \bf{0.00 $\pm$   0.00}\\
100&Eq+TD&  13.02 $\pm$   0.10& --&  58.34 $\pm$   2.51&  58.24 $\pm$   2.50&   \bf{0.10 $\pm$   0.05}&   \bf{0.00 $\pm$   0.00}\\
100&NoCurl-1{\color{red}s}&   6.55 $\pm$   0.30& 100.83 $\pm$   0.84&  57.03 $\pm$   1.49&  38.63 $\pm$   1.20&  14.85 $\pm$   0.38&   3.55 $\pm$   0.20\\
100&NoCurl-2{\color{red}s}&  21.11 $\pm$   0.56&  66.03 $\pm$   0.66&  35.75 $\pm$   1.17&  25.47 $\pm$   0.95&   7.96 $\pm$   0.29&   2.32 $\pm$   0.15\\
100&NoCurl-1&  18.20 $\pm$   0.81&  -0.82 $\pm$   0.16&  31.99 $\pm$   1.66&  24.09 $\pm$   1.32&   1.30 $\pm$   0.16&   6.60 $\pm$   0.30\\
100&NoCurl-2&  26.02 $\pm$   0.76&  -1.44 $\pm$   0.09&  \bf{19.16 $\pm$   1.10}&  14.30 $\pm$   0.89&   {0.62 $\pm$   0.09}&   4.24 $\pm$   0.23\\
\hline
\end{tabular}
\end{table*}
\begin{table*}[!ht]\renewcommand{\arraystretch}{0.9}
\centering
\caption{Comparison of Different Algorithms on Linear Synthetic Datasets: results (mean $\pm$ standard error over 100 trials) for ER4-Gaussian Cases, where bold numbers highlight the best method for each case.}
\label{Table:score2}
\vskip 0.1in
\small
\begin{tabular}{cccccccc}
\hline
$d$&Method&Time&$\Delta F$&SHD&\#Extra E&\#Missing E&\#Reverse E\\
\hline
10&NOTEARS&   3.35 $\pm$   0.13&   \bf{0.08 $\pm$   0.02}&   \bf{1.88 $\pm$   0.26}&   \bf{0.89 $\pm$   0.15}&   0.35 $\pm$   0.07&   {0.64 $\pm$   0.08}\\
10&FGS&   0.80 $\pm$   0.08&   --&  13.14 $\pm$   0.69&   6.88 $\pm$   0.43&   2.56 $\pm$   0.20&   3.70 $\pm$   0.24\\
10&CAM&  12.10 $\pm$   0.17&   --&  19.06 $\pm$   0.64&   7.51 $\pm$   0.31&   4.35 $\pm$   0.27&   7.20 $\pm$   0.28\\
10&MMPC&   1.14 $\pm$   0.04&   --&  21.13 $\pm$   0.39&   1.45 $\pm$   0.12&   9.16 $\pm$   0.41&  10.52 $\pm$   0.19\\
10&Eq+BU&   0.55 $\pm$   0.01& --&   4.73 $\pm$   0.24&   4.58 $\pm$   0.23&   0.15 $\pm$   0.06&   \bf{0.00 $\pm$   0.00}\\
10&Eq+TD&   0.55 $\pm$   0.01& --&   4.81 $\pm$   0.25&   4.67 $\pm$   0.23&   \bf{0.14 $\pm$   0.05}&   \bf{0.00 $\pm$   0.00}\\
10&NoCurl-1{\color{red}s}&   \bf{0.10 $\pm$   0.00}&   4.44 $\pm$   0.09&   4.01 $\pm$   0.36&   2.02 $\pm$   0.21&   1.50 $\pm$   0.16&   0.49 $\pm$   0.07\\
10&NoCurl-2{\color{red}s}&   0.56 $\pm$   0.02&   2.03 $\pm$   0.07&   2.39 $\pm$   0.29&   1.11 $\pm$   0.16&   0.57 $\pm$   0.14&   0.71 $\pm$   0.08\\
10&NoCurl-1&   {0.25 $\pm$   0.01}&   0.14 $\pm$   0.03&   2.51 $\pm$   0.32&   1.39 $\pm$   0.19&   0.45 $\pm$   0.09&   0.67 $\pm$   0.08\\
10&NoCurl-2&   0.43 $\pm$   0.01&   0.13 $\pm$   0.03&   2.22 $\pm$   0.27&   1.12 $\pm$   0.17&   {0.33 $\pm$   0.06}&   0.77 $\pm$   0.08\\

\hline
30&NOTEARS&  94.21 $\pm$   5.25&   \bf{0.25 $\pm$   0.11}&   8.81 $\pm$   1.08&   6.11 $\pm$   0.78&   1.50 $\pm$   0.28&   {1.20 $\pm$   0.11}\\
30&FGS&   1.71 $\pm$   0.10&   --&  50.37 $\pm$   3.27&  37.87 $\pm$   2.75&   5.50 $\pm$   0.42&   7.00 $\pm$   0.31\\
30&CAM&  61.92 $\pm$   0.78&   --&  56.80 $\pm$   1.69&  29.98 $\pm$   0.95&  11.96 $\pm$   0.61&  14.86 $\pm$   0.39\\
30&MMPC&   1.58 $\pm$   0.05&   --&  63.70 $\pm$   0.80&   \bf{4.19 $\pm$   0.21}&  27.61 $\pm$   1.00&  31.90 $\pm$   0.47\\
30&Eq+BU&   2.42 $\pm$   0.04& --&  31.06 $\pm$   1.33&  30.79 $\pm$   1.31&   \bf{0.27 $\pm$   0.08}&   \bf{0.00 $\pm$   0.00}\\
30&Eq+TD&   2.36 $\pm$   0.04& --&  33.48 $\pm$   1.46&  33.16 $\pm$   1.43&   0.32 $\pm$   0.08&   \bf{0.00 $\pm$   0.00}\\
30&NoCurl-1{\color{red}s}&   \bf{0.59 $\pm$   0.03}&  58.05 $\pm$   1.08&  20.27 $\pm$   0.81&  13.17 $\pm$   0.61&   6.09 $\pm$   0.27&   1.01 $\pm$   0.09\\
30&NoCurl-2{\color{red}s}&   1.92 $\pm$   0.06&  37.25 $\pm$   0.68&  13.16 $\pm$   0.87&   8.67 $\pm$   0.65&   3.75 $\pm$   0.25&   0.74 $\pm$   0.08\\
30&NoCurl-1&   {1.18 $\pm$   0.06}&   0.31 $\pm$   0.05&  10.84 $\pm$   0.72&   7.97 $\pm$   0.56&   0.75 $\pm$   0.09&   2.12 $\pm$   0.13\\
30&NoCurl-2&   3.39 $\pm$   0.11&   0.40 $\pm$   0.19&   \bf{7.91 $\pm$   0.83}&   {5.69 $\pm$   0.68}&   {0.71 $\pm$   0.12}&   1.51 $\pm$   0.12\\

\hline
50&NOTEARS& 209.49 $\pm$   6.13&   \bf{0.19 $\pm$   0.09}&  19.98 $\pm$   1.46&  14.73 $\pm$   1.12&   3.45 $\pm$   0.37&   {1.80 $\pm$   0.14}\\
50&FGS&   {3.41 $\pm$   0.20}&   --&  71.11 $\pm$   4.15&  54.36 $\pm$   3.63&   7.81 $\pm$   0.44&   8.94 $\pm$   0.38\\
50&CAM& 117.45 $\pm$   1.92&   --&  91.13 $\pm$   2.01&  47.89 $\pm$   1.23&  18.91 $\pm$   0.73&  24.33 $\pm$   0.49\\
50&MMPC&   3.84 $\pm$   0.15&   --& 106.73 $\pm$   1.07&   \bf{6.07 $\pm$   0.28}&  44.99 $\pm$   1.20&  55.67 $\pm$   0.58\\
50&Eq+BU&   4.25 $\pm$   0.03& --&  64.18 $\pm$   2.29&  63.68 $\pm$   2.26&   \bf{0.50 $\pm$   0.12}&   \bf{0.00 $\pm$   0.00}\\
50&Eq+TD&   3.98 $\pm$   0.03& --&  69.71 $\pm$   2.40&  69.21 $\pm$   2.37&   \bf{0.50 $\pm$   0.12}&   \bf{0.00 $\pm$   0.00}\\
50&NoCurl-1{\color{red}s}&    \bf{3.33 $\pm$   0.17}& 164.65 $\pm$   2.42&  37.82 $\pm$   1.19&  25.68 $\pm$   0.94&  10.38 $\pm$   0.36&   1.76 $\pm$   0.14\\
50&NoCurl-2{\color{red}s}&   5.80 $\pm$   0.19& 110.84 $\pm$   1.86&  26.26 $\pm$   1.13&  18.35 $\pm$   0.89&   6.55 $\pm$   0.29&   1.36 $\pm$   0.12\\
50&NoCurl-1&   4.13 $\pm$   0.20&   0.40 $\pm$   0.13&  19.76 $\pm$   1.22&  15.01 $\pm$   1.01&   1.02 $\pm$   0.12&   3.73 $\pm$   0.19\\
50&NoCurl-2&   7.55 $\pm$   0.25&   0.42 $\pm$   0.21&  \bf{15.24 $\pm$   1.27}&  11.66 $\pm$   1.04&   {0.81 $\pm$   0.13}&   2.77 $\pm$   0.20\\
\hline
100&NOTEARS& 1265.47 $\pm$  15.70&  \bf{-0.64 $\pm$   0.14}&  49.07 $\pm$   2.55&  37.86 $\pm$   2.07&   8.27 $\pm$   0.49&   {2.94 $\pm$   0.18}\\
100&FGS&   \bf{10.17 $\pm$   0.65}&   --&  93.24 $\pm$   5.63&  66.74 $\pm$   4.67&  12.98 $\pm$   0.83&  13.52 $\pm$   0.46\\
100&CAM& 258.39 $\pm$   1.83&   --& 159.91 $\pm$   3.10&  81.10 $\pm$   1.83&  34.55 $\pm$   1.23&  44.26 $\pm$   0.72\\
100&MMPC&  14.10 $\pm$   0.56&   --& 213.12 $\pm$   1.49&  \bf{12.00 $\pm$   0.39}&  83.00 $\pm$   1.84& 118.12 $\pm$   1.14\\
100&Eq+BU&  15.21 $\pm$   0.19& --& 138.38 $\pm$   5.52& 137.61 $\pm$   5.47&   \bf{0.77 $\pm$   0.14}&   \bf{0.00 $\pm$   0.00}\\
100&Eq+TD&  12.69 $\pm$   0.15& --& 150.08 $\pm$   6.03& 149.31 $\pm$   5.97&   \bf{0.77 $\pm$   0.13}&   \bf{0.00 $\pm$   0.00}\\
100&NoCurl-1{\color{red}s}&  12.90 $\pm$   0.69& 492.97 $\pm$   7.35&  82.46 $\pm$   1.45&  58.14 $\pm$   1.25&  21.13 $\pm$   0.43&   3.19 $\pm$   0.17\\
100&NoCurl-2{\color{red}s}&  30.94 $\pm$   1.13& 348.21 $\pm$   5.30&  60.64 $\pm$   1.79&  43.99 $\pm$   1.51&  13.80 $\pm$   0.42&   2.85 $\pm$   0.18\\
100&NoCurl-1&  26.43 $\pm$   1.46&  -0.28 $\pm$   0.26&  44.43 $\pm$   1.80&  34.83 $\pm$   1.50&   1.74 $\pm$   0.17&   7.86 $\pm$   0.28\\
100&NoCurl-2&  43.99 $\pm$   1.77&  -0.32 $\pm$   0.28&  \bf{37.11 $\pm$   1.71}&  29.28 $\pm$   1.48&   {1.57 $\pm$   0.16}&   6.26 $\pm$   0.23\\
\hline
\end{tabular}
\end{table*}
\begin{table*}[!ht]\renewcommand{\arraystretch}{0.9}
\centering
\caption{Comparison of Different Algorithms on Linear Synthetic Datasets: results (mean $\pm$ standard error over 100 trials) for ER6-Gaussian Cases, where bold numbers highlight the best method for each case.}
\label{Table:score3}
\vskip 0.1in
\small
\begin{tabular}{cccccccc}
\hline
$d$&Method&Time&$\Delta F$&SHD&\#Extra E&\#Missing E&\#Reverse E\\
\hline
10&NOTEARS&   3.20 $\pm$   0.20&  \bf{ 0.22 $\pm$   0.04}&   3.21 $\pm$   0.31&   1.11 $\pm$   0.14&   \bf{1.00 $\pm$   0.14}&   1.10 $\pm$   0.09\\
10&FGS&   0.58 $\pm$   0.02&   --&  19.77 $\pm$   0.58&   8.19 $\pm$   0.28&   5.42 $\pm$   0.21&   6.16 $\pm$   0.35\\
10&CAM&  10.46 $\pm$   0.17&   --&  26.41 $\pm$   0.46&   6.07 $\pm$   0.24&  11.17 $\pm$   0.30&   9.17 $\pm$   0.31\\
10&MMPC&   1.14 $\pm$   0.04&   --&  21.13 $\pm$   0.39&   1.45 $\pm$   0.12&   9.16 $\pm$   0.41&  10.52 $\pm$   0.19\\
10&Eq+BU&   0.56 $\pm$   0.01& --&   6.18 $\pm$   0.31&   4.79 $\pm$   0.21&   1.39 $\pm$   0.22&   \bf{0.00 $\pm$   0.00}\\
10&Eq+TD&   0.57 $\pm$   0.01& --&   6.22 $\pm$   0.30&   4.85 $\pm$   0.20&   1.37 $\pm$   0.23&   \bf{0.00 $\pm$   0.00}\\
10&NoCurl-1{\color{red}s}&   0.37 $\pm$   0.03&   8.23 $\pm$   0.15&   4.79 $\pm$   0.37&   1.78 $\pm$   0.18&   2.32 $\pm$   0.23&   0.69 $\pm$   0.08\\
10&NoCurl-2{\color{red}s}&   0.92 $\pm$   0.03&   5.55 $\pm$   0.32&   3.27 $\pm$   0.35&   1.10 $\pm$   0.16&   1.19 $\pm$   0.17&   0.98 $\pm$   0.09\\
10&NoCurl-1&   \bf{0.34 $\pm$   0.02}&   0.54 $\pm$   0.22&   3.54 $\pm$   0.37&   1.37 $\pm$   0.18&   1.30 $\pm$   0.17&   {0.87 $\pm$   0.08}\\
10&NoCurl-2&   0.75 $\pm$   0.03&   0.36 $\pm$   0.07&   \bf{3.07 $\pm$   0.30}&   \bf{0.97 $\pm$   0.13}&   1.02 $\pm$   0.14&   1.08 $\pm$   0.09\\

\hline
30&NOTEARS& 102.46 $\pm$   4.68&   1.02 $\pm$   0.18&  20.85 $\pm$   2.09&  15.15 $\pm$   1.58&   3.75 $\pm$   0.49&   {1.95 $\pm$   0.14}\\
30&FGS&   5.44 $\pm$   0.21&   --& 132.42 $\pm$   3.71& 105.01 $\pm$   3.06&  14.64 $\pm$   0.52&  12.77 $\pm$   0.53\\
30&CAM&  64.53 $\pm$   0.75&   --& 105.49 $\pm$   1.86&  50.80 $\pm$   1.08&  35.69 $\pm$   0.90&  19.00 $\pm$   0.45\\
30&MMPC&   \bf{1.58 $\pm$   0.05}&   --&  63.70 $\pm$   0.80&   \bf{4.19 $\pm$   0.21}&  27.61 $\pm$   1.00&  31.90 $\pm$   0.47\\
30&Eq+BU&   2.12 $\pm$   0.03& --&  68.38 $\pm$   1.66&  63.70 $\pm$   1.40&   4.68 $\pm$   0.57&   \bf{0.00 $\pm$   0.00}\\
30&Eq+TD&   2.08 $\pm$   0.02& --&  70.82 $\pm$   1.67&  66.04 $\pm$   1.40&   4.78 $\pm$   0.57&   \bf{0.00 $\pm$   0.00}\\
30&NoCurl-1{\color{red}s}&   \bf{1.58 $\pm$   0.10}& 824.91 $\pm$  15.39&  37.36 $\pm$   1.34&  24.83 $\pm$   0.94&  11.04 $\pm$   0.47&   1.49 $\pm$   0.14\\
30&NoCurl-2{\color{red}s}&   4.69 $\pm$   0.23& 667.06 $\pm$  17.54&  29.88 $\pm$   1.55&  20.25 $\pm$   1.12&   8.01 $\pm$   0.46&   1.62 $\pm$   0.13\\
30&NoCurl-1&   3.34 $\pm$   0.21&   1.78 $\pm$   0.38&  21.44 $\pm$   1.56&  15.98 $\pm$   1.16&   2.41 $\pm$   0.33&   3.05 $\pm$   0.19\\
30&NoCurl-2&   7.68 $\pm$   0.39&   \bf{0.97 $\pm$   0.16}&  \bf{17.37 $\pm$   1.18}&  12.93 $\pm$   0.92&   \bf{1.71 $\pm$   0.19}&   2.73 $\pm$   0.16\\
\hline
50&NOTEARS& 340.03 $\pm$   6.99&   1.97 $\pm$   0.26&  52.40 $\pm$   3.24&  40.53 $\pm$   2.62&   9.25 $\pm$   0.67&   {2.62 $\pm$   0.16}\\
50&FGS&  20.31 $\pm$   1.00&   --& 235.85 $\pm$   7.94& 195.75 $\pm$   6.77&  23.79 $\pm$   1.00&  16.31 $\pm$   0.56\\
50&CAM& 129.50 $\pm$   1.18&   --& 176.25 $\pm$   2.94&  89.77 $\pm$   1.73&  59.87 $\pm$   1.27&  26.61 $\pm$   0.53\\
50&MMPC&   \bf{3.84 $\pm$   0.15}&   --& 106.73 $\pm$   1.07&   \bf{6.07 $\pm$   0.28}&  44.99 $\pm$   1.20&  55.67 $\pm$   0.58\\
50&Eq+BU&   3.97 $\pm$   0.03& --& 153.43 $\pm$   3.76& 145.16 $\pm$   3.27&   8.27 $\pm$   0.93&   \bf{0.00 $\pm$   0.00}\\
50&Eq+TD&   \bf{3.84 $\pm$   0.02}& --& 161.11 $\pm$   4.00& 152.68 $\pm$   3.52&   8.43 $\pm$   0.97&   \bf{0.00 $\pm$   0.00}\\
50&NoCurl-1{\color{red}s}&   8.45 $\pm$   0.55& 2957.04 $\pm$  69.41&  69.24 $\pm$   2.00&  48.31 $\pm$   1.57&  19.01 $\pm$   0.54&   1.92 $\pm$   0.15\\
50&NoCurl-2{\color{red}s}&  12.51 $\pm$   0.68& 2114.01 $\pm$  51.61&  56.68 $\pm$   1.95&  40.93 $\pm$   1.55&  13.62 $\pm$   0.52&   2.13 $\pm$   0.14\\
50&NoCurl-1&  12.05 $\pm$   0.77&   2.31 $\pm$   0.41&  40.32 $\pm$   2.40&  32.10 $\pm$   2.00&   2.83 $\pm$   0.24&   5.39 $\pm$   0.26\\
50&NoCurl-2&  31.74 $\pm$   1.71&   \bf{1.77 $\pm$   0.38}&  \bf{33.67 $\pm$   2.53}&  26.69 $\pm$   2.08&   \bf{2.45 $\pm$   0.35}&   {4.53 $\pm$   0.22}\\
\hline
100&NOTEARS& 2146.90 $\pm$  31.22&   \bf{2.49 $\pm$   0.37}& 116.52 $\pm$   4.39&  92.10 $\pm$   3.66&  20.54 $\pm$   0.84&   {3.88 $\pm$   0.21}\\
100&FGS& 105.57 $\pm$   5.35&   --& 421.53 $\pm$  15.01& 356.15 $\pm$  13.33&  43.64 $\pm$   1.59&  21.74 $\pm$   0.62\\
100&CAM& 290.21 $\pm$   3.76&   --& 310.54 $\pm$   4.54& 156.83 $\pm$   2.66& 110.55 $\pm$   2.03&  43.16 $\pm$   0.69\\
100&MMPC&  {14.10 $\pm$   0.56}&   --& 213.12 $\pm$   1.49&  \bf{12.00 $\pm$   0.39}&  83.00 $\pm$   1.84& 118.12 $\pm$   1.14\\
100&Eq+BU&  13.15 $\pm$   0.15& --& 378.33 $\pm$   8.50& 365.13 $\pm$   7.81&  13.20 $\pm$   1.20&   \bf{0.00 $\pm$   0.00}\\
100&Eq+TD&  \bf{11.37 $\pm$   0.11}& --& 397.65 $\pm$   8.66& 383.96 $\pm$   7.92&  13.69 $\pm$   1.20&   \bf{0.00 $\pm$   0.00}\\
100&NoCurl-1{\color{red}s}&  43.98 $\pm$   2.11& 6319.57 $\pm$ 103.73& 138.70 $\pm$   2.61&  97.83 $\pm$   2.14&  37.22 $\pm$   0.69&   3.65 $\pm$   0.16\\
100&NoCurl-2{\color{red}s}& 106.88 $\pm$   4.64& 5285.84 $\pm$ 124.45& 108.05 $\pm$   2.74&  78.33 $\pm$   2.27&  26.03 $\pm$   0.60&   3.69 $\pm$   0.19\\
100&NoCurl-1&  41.29 $\pm$   1.56&   4.30 $\pm$   0.99&  89.93 $\pm$   3.49&  74.34 $\pm$   3.04&   4.64 $\pm$   0.34&  10.95 $\pm$   0.33\\
100&NoCurl-2&  84.24 $\pm$   3.26&   2.61 $\pm$   0.93&  \bf{72.30 $\pm$   3.80}&  60.12 $\pm$   3.49&   \bf{3.58 $\pm$   0.30}&   8.60 $\pm$   0.27\\
\hline
\end{tabular}
\end{table*}

\begin{table*}[!ht]\renewcommand{\arraystretch}{0.9}
\centering
\caption{Comparison of Different Algorithms on Linear Synthetic Datasets: results (mean $\pm$ standard error over 100 trials) for SF4-Gumbel Cases, where bold numbers highlight the best method for each case.}
\label{Table:score4}
\vskip 0.1in
\small
\begin{tabular}{cccccccc}
\hline
$d$&Method&Time&$\Delta F$&SHD&\#Extra E&\#Missing E&\#Reverse E\\
\hline
10&NOTEAR&   5.26 $\pm$   0.17&  \bf{-0.71 $\pm$   0.01}&   1.10 $\pm$   0.22&   0.80 $\pm$   0.15&   0.12 $\pm$   0.05&   0.18 $\pm$   0.04\\
10&FGS&   0.47 $\pm$   0.02&   --&   5.30 $\pm$   0.57&   3.13 $\pm$   0.44&   0.99 $\pm$   0.12&   1.18 $\pm$   0.11\\
10&CAM&  11.76 $\pm$   0.20&   --&  17.70 $\pm$   0.73&   9.22 $\pm$   0.49&   1.67 $\pm$   0.13&   6.81 $\pm$   0.27\\
10&MMPC&   0.52 $\pm$   0.02&   --&  14.93 $\pm$   0.18&   0.88 $\pm$   0.11&   3.06 $\pm$   0.17&  10.99 $\pm$   0.13\\
10&Eq+BU&   0.67 $\pm$   0.01& --&   1.24 $\pm$   0.13&   1.24 $\pm$   0.13&   \bf{0.00 $\pm$   0.00}&   \bf{0.00 $\pm$   0.00}\\
10&Eq+TD&   0.55 $\pm$   0.02& --&   1.28 $\pm$   0.13&   1.27 $\pm$   0.13&   0.01 $\pm$   0.01&   \bf{0.00 $\pm$   0.00}\\
10&NoCurl-1{\color{red}s}&   \bf{0.06 $\pm$   0.00}&   0.09 $\pm$   0.01&   0.94 $\pm$   0.17&   \bf{0.67 $\pm$   0.13}&   0.16 $\pm$   0.04&   0.11 $\pm$   0.03\\
10&NoCurl-2{\color{red}s}&   0.18 $\pm$   0.00&  -0.11 $\pm$   0.00&   0.97 $\pm$   0.17&   0.73 $\pm$   0.13&   0.03 $\pm$   0.02&   0.21 $\pm$   0.05\\
10&NoCurl-1&   {0.14 $\pm$   0.00}&  -0.58 $\pm$   0.01&   \bf{0.93 $\pm$   0.20}&   {0.69 $\pm$   0.15}&   0.08 $\pm$   0.04&   {0.16 $\pm$   0.04}\\
10&NoCurl-2&   0.35 $\pm$   0.01&  -0.59 $\pm$   0.01&   1.08 $\pm$   0.22&   0.86 $\pm$   0.19&   {0.04 $\pm$   0.02}&   0.18 $\pm$   0.04\\
\hline
30&NOTEARS&  82.37 $\pm$   1.57&  \bf{-3.55 $\pm$   0.02}&   2.68 $\pm$   0.51&   2.13 $\pm$   0.43&   0.11 $\pm$   0.05&   0.44 $\pm$   0.07\\
30&FGS&   1.01 $\pm$   0.03&   --&  22.81 $\pm$   1.70&  12.18 $\pm$   1.38&   7.68 $\pm$   0.46&   2.95 $\pm$   0.19\\
30&CAM&  62.27 $\pm$   0.91&   --&  62.80 $\pm$   1.29&  28.46 $\pm$   0.96&  13.71 $\pm$   0.32&  20.63 $\pm$   0.42\\
30&MMPC&  13.58 $\pm$   3.40&   --&  54.24 $\pm$   0.39&   4.14 $\pm$   0.23&  18.61 $\pm$   0.42&  31.49 $\pm$   0.34\\
30&Eq+BU&   2.74 $\pm$   0.05& --&   9.46 $\pm$   0.49&   9.42 $\pm$   0.49&   \bf{0.04 $\pm$   0.02}&   \bf{0.00 $\pm$   0.00}\\
30&Eq+TD&   3.00 $\pm$   0.03& --&   9.91 $\pm$   0.52&   9.86 $\pm$   0.52&   0.05 $\pm$   0.02&   \bf{0.00 $\pm$   0.00}\\
30&NoCurl-1{\color{red}s}&   \bf{0.29 $\pm$   0.01}&   5.57 $\pm$   0.08&   5.76 $\pm$   0.59&   4.85 $\pm$   0.52&   0.67 $\pm$   0.10&   0.24 $\pm$   0.05\\
30&NoCurl-2{\color{red}s}&   1.13 $\pm$   0.02&   2.28 $\pm$   0.07&   5.26 $\pm$   0.75&   4.35 $\pm$   0.65&   0.37 $\pm$   0.10&   0.54 $\pm$   0.07\\
30&NoCurl-1&   {0.76 $\pm$   0.02}&  -3.30 $\pm$   0.04&   \bf{2.57 $\pm$   0.43}&   \bf{2.04 $\pm$   0.37}&   0.12 $\pm$   0.04&   {0.41 $\pm$   0.05}\\
30&NoCurl-2&   1.84 $\pm$   0.05&  -3.31 $\pm$   0.02&   4.42 $\pm$   0.70&   3.60 $\pm$   0.62&   {0.09 $\pm$   0.03}&   0.73 $\pm$   0.09\\
\hline
50&NOTEARS& 150.33 $\pm$   1.98&  \bf{-7.08 $\pm$   0.02}&   \bf{3.94 $\pm$   0.77}&   3.22 $\pm$   0.70&   0.18 $\pm$   0.07&   {0.54 $\pm$   0.07}\\
50&FGS&   {2.35 $\pm$   0.10}&   --&  43.47 $\pm$   2.77&  19.25 $\pm$   2.12&  19.00 $\pm$   0.95&   5.22 $\pm$   0.29\\
50&CAM& 110.65 $\pm$   1.42&   --& 103.32 $\pm$   1.55&  39.74 $\pm$   1.10&  30.54 $\pm$   0.69&  33.04 $\pm$   0.52\\
50&MMPC& 417.94 $\pm$ 349.20&   --&  96.70 $\pm$   0.56&   8.91 $\pm$   0.38&  38.39 $\pm$   0.83&  49.40 $\pm$   0.69\\
50&Eq+BU&   5.49 $\pm$   0.05& --&  23.64 $\pm$   1.03&  23.30 $\pm$   1.02&   0.33 $\pm$   0.12&   0.01 $\pm$   0.01\\
50&Eq+TD&   5.75 $\pm$   0.04& --&  24.52 $\pm$   1.08&  24.18 $\pm$   1.07&   0.34 $\pm$   0.12&   \bf{0.00 $\pm$   0.00}\\
50&NoCurl-1{\color{red}s}&   \bf{0.99 $\pm$   0.05}&  24.11 $\pm$   0.37&  14.98 $\pm$   1.13&  12.82 $\pm$   1.01&   1.72 $\pm$   0.15&   0.44 $\pm$   0.08\\
50&NoCurl-2{\color{red}s}&   5.94 $\pm$   0.15&  10.42 $\pm$   0.21&  13.73 $\pm$   1.36&  11.78 $\pm$   1.24&   0.56 $\pm$   0.08&   1.39 $\pm$   0.12\\
50&NoCurl-1&   3.45 $\pm$   0.13&  -6.74 $\pm$   0.03&   4.06 $\pm$   0.64&   \bf{3.16 $\pm$   0.56}&   \bf{0.14 $\pm$   0.03}&   0.76 $\pm$   0.09\\
50&NoCurl-2&   5.64 $\pm$   0.14&  -6.74 $\pm$   0.03&   8.38 $\pm$   1.17&   7.05 $\pm$   1.08&   0.18 $\pm$   0.05&   1.15 $\pm$   0.10\\
\hline
100&NOTEARS& 1113.10 $\pm$   9.71& \bf{-17.53 $\pm$   0.05}&  11.98 $\pm$   2.18&  10.40 $\pm$   2.04&   0.43 $\pm$   0.11&   {1.15 $\pm$   0.12}\\
100&FGS&   {8.04 $\pm$   0.54}&   --&  91.32 $\pm$   3.48&  30.09 $\pm$   2.59&  52.39 $\pm$   1.54&   8.84 $\pm$   0.34\\
100&CAM& 240.04 $\pm$   2.91&   --& 211.33 $\pm$   2.25&  74.66 $\pm$   1.60&  76.12 $\pm$   0.90&  60.55 $\pm$   0.82\\
100&MMPC&  40.22 $\pm$  14.80&   --& 217.00 $\pm$   0.82&  32.41 $\pm$   0.73&  88.73 $\pm$   1.12&  95.86 $\pm$   1.04\\
100&Eq+BU&  21.50 $\pm$   0.29& --&  62.96 $\pm$   2.20&  61.33 $\pm$   2.27&   1.62 $\pm$   0.30&   0.01 $\pm$   0.01\\
100&Eq+TD&  17.46 $\pm$   0.10& --&  65.60 $\pm$   2.27&  63.98 $\pm$   2.34&   1.62 $\pm$   0.30&   \bf{0.00 $\pm$   0.00}\\
100&NoCurl-1{\color{red}s}&  \bf{6.87 $\pm$   0.33}&  97.63 $\pm$   1.31&  38.66 $\pm$   2.28&  35.02 $\pm$   2.09&   3.03 $\pm$   0.25&   0.61 $\pm$   0.09\\
100&NoCurl-2{\color{red}s}&  23.98 $\pm$   0.87&  55.31 $\pm$   1.21&  30.14 $\pm$   2.42&  26.66 $\pm$   2.23&   1.78 $\pm$   0.20&   1.70 $\pm$   0.15\\
100&NoCurl-1&  27.64 $\pm$   0.82& -17.29 $\pm$   0.05&   \bf{8.68 $\pm$   1.09}&   \bf{7.06 $\pm$   1.01}&   0.19 $\pm$   0.04&   1.43 $\pm$   0.13\\
100&NoCurl-2&  49.83 $\pm$   1.20& -17.19 $\pm$   0.06&  16.84 $\pm$   1.66&  14.78 $\pm$   1.58&   \bf{0.17 $\pm$   0.05}&   1.89 $\pm$   0.14\\
\hline
\end{tabular}
\end{table*}

\section{Detailed Results for Nonlinear Synthetic Datasets}\label{sec:nonlinear}

In this section we provide additional results and discussions for the experiments on nonlinear SEM datasets in Section \ref{sec:data2} of the main text, with the details of dataset generation and algorithm settings provided in  Section \ref{sec:nonlinearsetting} above. 
Table~\ref{Table:gnn} shows the full results of the various methods in nonlinear synthetic datasets. We note that some $d=100$ results for NOTEARS-MLP, GraN-DAG, and GSGES are missing since these algorithms could not finish within 72 hours on at least one trial.\footnote{Although GraN-DAG without the pruning phase finishes within 72 hours in Nonlinear Case 1 and Nonlinear Case 3, the results are with a very large SHD ($762\pm345$ in Nonlinear Case 1 and $1606\pm421$ in Nonlinear Case 3), since the pruning step generally has the effect of greatly reducing the SHD, as reported in \citep{lachapelle2019gradient}.} Moreover, we also observe that the run time for MMPC vary drastically, potentially due to the conditioned variable set size. When its size is large, exhaustive search becomes prohibitively expensive. CAM is the most accurate non-neural-network methods followed by GSGES. 
However, GSGES's running time get prohibitive with large dimensions. 

As one can see, in the Nonlinear Case 1 datasets, GraN-DAG and CAM perform the best among all the methods but become the worst in Nonlinear Case 3 datasets among finished methods.  NOTEARS-MLP performs the best in Nonlinear Case 3 but does not do as well in Nonlinear Case 1 and 2 and have trouble handling larger graphs. DAG-GNN and NoCurl does the best in Nonlinear Case 2 cases in comparison and are able to beat NOTEARS-MLP in Nonlinear Case 1 when d is larger. It shows there is no universal best nonlinear DAG learner. Moreover, NoCurl with DAG-GNN as the base model performs as well as DAG-GNN and takes about 3 to 4 times faster, which indicate that NoCurl has successfully boost the efficiency of DAG-GNN without deteriorate its accuracy. NoCurl is also more than one order of magnitude of faster than GraN-DAG  and NOTEARS-MLP in many testing cases. 

As discussed in the main text, it should be interesting to extend NoCurl to use NOTEARS-MLP and GraN-DAG as base models by considering gradient-based adjacency matrix. 


\begin{table*}[!ht]\renewcommand{\arraystretch}{0.9}
\centering
\caption{Comparison of Different Algorithms on Nonlinear Synthetic datasets: results (mean $\pm$ standard error over $5$ trails) on SHD and Run Time (in seconds), where bold numbers highlight the best method for each case.}
\label{Table:gnn}
\vskip 0.1in
\scriptsize
\begin{tabular}{lccccc cc}
\hline
\hline
\multicolumn{7}{c}{\textbf{Nonlinear Case 1: SHD}}\\
 d &  NOTEARS-MLP & GraN-DAG & CAM & MMPC &GSGES & DAG-GNN & DAG-GNN + NoCurl \\
\hline
$10$& $\bf{3.0\pm1.2}$ & $3.2\pm1.7$ & $4.6\pm0.6$ &$21.4 \pm0.4$ &$9.0 \pm3.9$ & $7.4\pm2.8 $& $7.4\pm3.2 $\\
$20$& $\bf{3.8\pm2.3}$ & $5.0\pm1.8$ & $12.6\pm1.5$ &$46.4\pm0.7$ &$12.6 \pm3.8$ & $8.8\pm3.1$& $8.6\pm3.5 $\\
$50$&  $29.0\pm6.5$ & $\bf{9.6\pm1.6}$ & $12.0\pm1.4$ &$110.6\pm2.7$  &$28.0 \pm3.7$ & $26.4\pm10.5$& $23.2\pm9.8 $\\
$100$& $>72h$ & $>72h$ & $\bf{34.2\pm4.0}$ &$251.2\pm6.1$  &$>72h$ & $58.6\pm15.9 $& $54.0\pm13.8$\\
\hline
\multicolumn{7}{c}{\textbf{Nonlinear Case 2: SHD}}\\
d&  NOTEARS-MLP & GraN-DAG & CAM & MMPC&GSGES & DAG-GNN & DAG-GNN + NoCurl \\
\hline
$10$& $\bf{0.4\pm0.4}$ & $4.2\pm2.4$ & $7.0\pm0.5$ &$15.8\pm0.3$  &$7.8 \pm4.7$ & $2.0\pm1.5$& $5.6\pm3.1$\\
$20$&$2.2\pm1.0$ & $8.0\pm3.1$ & $24.6\pm1.0$ &$37.0\pm0.3$  &$21.2 \pm11.8$ & $5.0\pm2.3$& $\bf{2.0\pm0.8}$\\
$50$&$20.8\pm4.7$ & $17.6\pm5.8$ & $41.4\pm1.2$ &$83.6\pm0.4$  &$54.6 \pm10.2$ & $12.4\pm4.6$& $\bf{9.0\pm3.4}$\\
$100$&$>72h$ & $26.0\pm6.7$ & $78.0\pm2.4$ &$179.0\pm1.5$  &$>72h$ & $21.4\pm2.3$& $\bf{18.6\pm3.0}$\\
\hline
  
\multicolumn{7}{c}{\textbf{Nonlinear Case 3: SHD}}\\
d &  NOTEARS-MLP & GraN-DAG & CAM & MMPC &GSGES& DAG-GNN & DAG-GNN + NoCurl \\
\hline
$10$&$2.6\pm1.2$ & $8.8\pm2.9$ & $0.2\pm0.0$ &$15.8\pm0.1 $  &$4.2 \pm0.8$ & $\bf{2.4\pm0.8}$& $3.0\pm0.6$\\
$20$&$\bf{6.0\pm2.5}$ & $36.0\pm11.1$&$13.6\pm1.0$  & $37.2\pm0.3$  &$18.8 \pm4.7$ & $9.0\pm1.4$& $10.4\pm1.6$\\
$50$&$\bf{14.8\pm0.4}$ & $60.8\pm8.0$ &$55.2\pm2.1$  & $>72h$  &$57.4 \pm5.2$ & $27.6\pm3.7$& $26.6\pm1.8$\\
$100$&$\bf{45.4\pm2.7}$ & $>72h$ &$73.0\pm0.5$ & $>72h$  &$>72h$ & $62.6\pm6.6$& $60.8\pm3.8$\\
\hline
\hline
\multicolumn{7}{c}{\textbf{Nonlinear Case 1: Run Time}}\\
d &  NOTEARS-MLP & GraN-DAG & CAM & MMPC&GSGES & DAG-GNN & DAG-GNN + NoCurl \\
\hline
$10$&$2.3e3\pm4.5e2$ & $7.2e2\pm5.2e1$ & $3.9e1\pm0.8$ &$\bf{1.5\pm0.1}$  &$6.3e2 \pm6.5e1$ & $6.3e1\pm1.4e2$& $3.8e2\pm3.3e1$\\
$20$&$9.1e3\pm15.3e4$ & $1.5e3\pm7.6e1$ & $8.9e1\pm1.2$ &$\bf{2.1\pm0.1}$  &$1.9e3 \pm3.2e2$
& $1.1e3\pm2.0e2$& $4.9e2\pm2.9e1$\\
$50$&$6.0e4\pm3.0e4$ & $5.3e3\pm4.2e2$ & $2.5e2\pm1.9$ &$\bf{1.3e1\pm1.4}$  &$1.7e4 \pm 1.1e3$ & $2.2e3\pm2.1e2$& $9.1e2\pm8.5e1$\\
$100$&$>72h$ & $>72h$ & $6.1e2\pm4.7$ &$\bf{1.0e2\pm7.6}$  &$>72h$ & $4.7e3\pm4.9e2$& $1.6e3\pm7.4e1$\\
\hline
\multicolumn{7}{c}{\textbf{Nonlinear Case 2: Run Time}}\\
d  &  NOTEARS-MLP & GraN-DAG & CAM & MMPC&GSGES & DAG-GNN & DAG-GNN + NoCurl \\
\hline
$10$&$3.2e3\pm9.4e2$ & $5.9e2\pm7.8e1$ & $3.5e1\pm1.2$ &$\bf{0.7\pm0.0}$  &$6.7e2 \pm3.8e1$
& $9.5e2\pm5.4e1$& $3.1e2\pm1.3e1$\\
$20$&$2.0e4\pm1.5e3$ & $1.3e3\pm5.3e2$ & $1.0e2\pm2.3$ &$\bf{1.1\pm0.0}$  &$1.9e3 \pm1.9e2$ & $1.2e3\pm7.2e1$& $3.9e2\pm1.3e1$\\
$50$&$1.8e5\pm5.4e4$ & $5.3e3\pm1.5e3$ & $2.9e2\pm4.6$ &$\bf{4.8\pm0.2}$  &$1.2e4 \pm1.4e3$ & $2.9e3\pm2.6e2$& $8.5e2\pm4.1e1$\\
$100$&$>72h$ & $1.6e4\pm7.2e2$ & $6.1e2\pm2.7$ &$\bf{1.1e1\pm0.4}$  &$>72h$ & $5.7e3\pm5.3e2$& $1.3e3\pm6.6e1$\\
\hline
\multicolumn{7}{c}{\textbf{Nonlinear Case 3: Run Time}}\\
d &  NOTEARS-MLP & GraN-DAG & CAM & MMPC&GSGES & DAG-GNN & DAG-GNN + NoCurl \\
\hline
$10$&$1.1e3\pm6.3e2$ & $1.3e3\pm2.1e2$& $ 5.1e1\pm0.7$  &$\bf{0.7\pm0.0}$  &$3.3e2 \pm1.4e1$ & $1.5e2\pm1.9e1$& $3.6e2\pm3.2e1$\\
$20$&$1.2e4\pm1.0e4$ & $2.2e3\pm4.4e2$ & $ 1.4e2\pm3.4$ &$\bf{4.5e1\pm6.7}$  &$1.6e3 \pm1.7e2$
& $1.3e3\pm1.3e2$& $4.2e2\pm4.1e1$\\
$50$&$7.8e3\pm1.1e3$ & $2.0e4\pm2.2e3$ & $\bf{3.8e2\pm7.3}$ &$>72h$  &$1.1e4 \pm 7.3e2$ & $2.7e3\pm3.2e2$& $9.9e2\pm5.2e1$\\
$100$&$3.2e4\pm6.2e3$ & $>72h$ & $\bf{7.9e2\pm5.5}$ &$>72h$  &$>72h$ & $4.7e3\pm3.5e2$& $1.3e3\pm3.1e1$\\
\hline
\hline
\end{tabular}
\end{table*}

\section{Learnt Protein Network}\label{sec:protein}

\begin{figure}[!ht]
\begin{center}
\subfigure{{\includegraphics[width=0.5\textwidth]{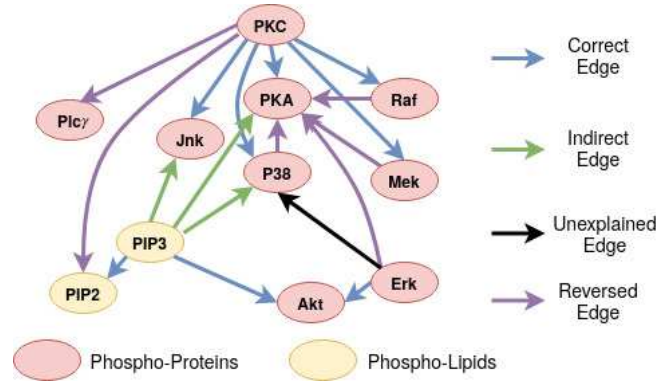}}}
\end{center}
  \caption{Learned protein signaling network.}
  \label{fig:protein}
\end{figure}

\begin{table*}[!ht]\renewcommand{\arraystretch}{0.99}
\centering
\caption{Accuracy Results on Protein Signaling Network, where bold number highlights the best method.} \label{Table:protein}
\small
\begin{tabular}{cccccccc}
\hline
Method & FGS & NOTEARS &NOTEARS-MLP&DAG-GNN & GraN-DAG&  CAM &DAG-GNN+NoCurl\\
\hline
\# Edges & 17&16&13 &18 & - & - &18 \\
SHD & 22 &22&16 &19 &  13$^*$ & \bf{12}$^*$ & 16 \\
\hline
\end{tabular}
\\
CAM and GraN-DAG results adopted from \citep{lachapelle2019gradient}, without the number of edges reported.
\end{table*}

We  now consider a real-world bioinformatics dataset~\citep{sachs2005causal} for the discovery of a protein signaling network based on expression levels of proteins and phospholipids. This is a widely used dataset for research on graphical models, with experimental annotations accepted by the biological research community. In Table~\ref{Table:protein}, we compare our results and 6 baseline methods against the ground truth offered in \citep{sachs2005causal}. On this dataset, DAG-NoCurl successfully learns the existence of 14 out of 20 ground-truth edges, and predicts the directions of 8 edges correctly (the learnt graph is plotted in Appendix Figure \ref{fig:protein}).


\end{document}